\documentclass[11pt,draftcls,onecolumn]{IEEEtran}

\usepackage{cite}

\usepackage{colortbl}
\usepackage{color}
\usepackage{bm}
\usepackage{bbm}
\usepackage{amsmath}

\usepackage{cases}

\usepackage{amssymb}
\usepackage{amsthm}
\usepackage{graphicx}
\usepackage{mathrsfs}
\usepackage{empheq}	
\usepackage[mathcal]{euscript}
\usepackage[margin = 2cm]{geometry}
\usepackage{framed}
\usepackage{comment}

\usepackage{dsfont}

\DeclareFontFamily{U}{mathx}{\hyphenchar\font45}
\DeclareFontShape{U}{mathx}{m}{n}{
      <5> <6> <7> <8> <9> <10>
      <10.95> <12> <14.4> <17.28> <20.74> <24.88>
      mathx10
      }{}
\DeclareSymbolFont{mathx}{U}{mathx}{m}{n}
\DeclareFontSubstitution{U}{mathx}{m}{n}
\DeclareMathAccent{\widecheck}{0}{mathx}{"71}

\newtheorem{theorem}{Theorem}
\newtheorem{lemma}{Lemma}

\newtheorem{assumption}{Assumption}

\newtheorem{property}{Property}

\newtheorem{example}{Example}

\newcommand{\E}{\mathbb{E}}

\newcommand{\beq}{\begin{equation}}
\newcommand{\eeq}{\end{equation}}
\newcommand{\beqa}{\begin{IEEEeqnarray}{rCl}}
\newcommand{\eeqa}{\end{IEEEeqnarray}}

\newcommand{\dfz}{\triangleq}

\title{Compressed Regression over Adaptive Networks}
\author{Marco Carpentiero, Vincenzo Matta, and Ali H. Sayed
\thanks{
Marco Carpentiero and Vincenzo Matta are with the Department of Information and Electrical Engineering and Applied Mathematics (DIEM), University of Salerno, via Giovanni Paolo II, I-84084, Fisciano (SA), Italy, and also with the National Inter-University Consortium for Telecommunications (CNIT), Italy (e-mails: \{mcarpentiero, vmatta\}@unisa.it).

Ali H. Sayed is with the \'Ecole Polytechnique F\'ed\'erale de Lausanne EPFL, School of Engineering, CH-1015 Lausanne, Switzerland (e-mail: ali.sayed@epfl.ch).
}
\thanks{A short conference version of this work will be presented at ICASSP 2023~\cite{CarpentieroMattaSayedICASSP2023}.}
}

\begin{document}
	\maketitle

\begin{abstract}
In this work we derive the performance achievable by a network of distributed agents that solve, {\em adaptively} and in the presence of {\em communication constraints}, a regression problem. Agents employ the recently proposed ACTC (adapt-compress-then-combine)  diffusion strategy, where the signals exchanged locally by neighboring agents are encoded with {\em randomized differential compression} operators. We provide a detailed characterization of the mean-square estimation error, which is shown to comprise a term related to the error that agents would achieve without communication constraints, plus a term arising from compression. The analysis reveals quantitative relationships between the compression loss and fundamental attributes of the distributed regression problem, in particular, the stochastic approximation error caused by the gradient noise and the network topology (through the Perron eigenvector). We show that knowledge of such relationships is critical to allocate optimally the communication resources across the agents, taking into account their individual attributes, such as the quality of their data or their degree of centrality in the network topology. 
We devise an optimized allocation strategy where the parameters necessary for the optimization can be learned {\em online} by the agents.
Illustrative examples show that a significant performance improvement, as compared to a blind (i.e., uniform) resource allocation, can be achieved by optimizing the allocation by means of the provided mean-square-error formulas.
\end{abstract}


\section{Introduction}
Motivated by the exponential growth of data availability and the success of networked device systems (e.g., Internet of Things, mobile edge computing, vehicular networks) {\em fully-decentralized strategies} and {\em federated strategies} \cite{McMahanMooreRamageAISTATS2017,LiSahuTalwalkarSIGMAG2020,Sayed,SayedTuChenZhaoTowficSPmag2013,SayedProcIEEE2014,PreddKulkarniPoorSPM2006, MattaSayedCoopGraphSP2018} represent the backbone of the next-generation learning algorithms. 
In this work we focus on the fully-decentralized scenario in the {\em adaptive} setting \cite{Sayed,SayedTuChenZhaoTowficSPmag2013,SayedProcIEEE2014}. We consider an online data-analysis problem where multi-agent networks, by means of {\em local interactions}, cooperatively learn an unknown model by observing streaming data. 
These types of problems can be successfully solved by combining popular learning algorithms, such as the {\em stochastic gradient algorithm} (with constant step-size, to track promptly data drifts), and distributed cooperation strategies like {\em consensus} \cite{DeGroot, XiaoBoydSCL2004, BoydGhoshPrabhakarShahTIT2006,DimakisKarMouraRabbatScaglioneProcIEEE2010} or {\em diffusion} \cite{Sayed,SayedTuChenZhaoTowficSPmag2013,SayedProcIEEE2014}. 

One of the biggest challenges of decentralized learning is the communication bottleneck induced by the back-and-forth information transmission among agents. Suitable compression strategies, e.g., quantization \cite{GershoGrayBook}, must be introduced to cope with unavoidable communication constraints. Data compression is a well-studied discipline, providing several useful tools that, over the years, have been successfully applied to many inference problems, even in distributed settings~\cite{LamReibmanTC1993,GubnerTIT1993,AsymptoticDesignQuantizers2007,LongoLookabaughGrayTIT1990,SaligramaTSP2006,HanAmariTIT1998}. However, online optimization problems such as the one addressed in this work pose additional challenges. First, the data distributions depend on an unknown parameter (possibly varying over time) that is the object of the optimization procedure, thus preventing the application of standard quantizer design methods. Second, because of the iterative nature of gradient-type algorithms, the distortion caused by quantization can accumulate over time, impairing the convergence of the optimization algorithm. Recent works showed that these issues can be addressed by means of {\em randomized} and/or {\em differential} compression~\cite{AlistarhNIPS2017,StichCordonnierJaggiNIPS2018,WangniWangLiuZhangNIPS2018,LinKostinaHassibiISIT2021}.

Randomized compression operators date back to the pioneering works on the probabilistic analysis of quantizers \cite{Widrow1956} and dithering \cite{GrayStockhamTIT1993}. By generating the coded output through suitable random mechanisms, these compression operators exhibit features useful for inference purposes, such as unbiasedness or mean-square-error boundedness, which hold  {\em universally}, i.e., irrespective of the particular data distribution. Thanks to these properties, the randomized compression approach has been recently employed in the distributed optimization field starting from the popular family of randomized quantizers \cite{AlistarhNIPS2017} and randomized sparsifiers \cite{StichCordonnierJaggiNIPS2018,WangniWangLiuZhangNIPS2018}.

Differential compression exploits techniques belonging to the legacy of communication systems, such as DPCM (Differential Pulse Code Modulation) and Delta Modulation \cite{GershoGrayBook}, to reduce the actual range of the encoder input by considering only the {\em innovation} between consecutive samples. This approach exploits the correlation existing between consecutive data samples to avoid wasting resources in the representation of redundant information. 

Several recent works have considered gradient descent and stochastic gradient descent algorithms in the presence of randomized compression~\cite{WuHuangHuangZhangICMLR2018,DoanMaguluriRombergIEEEAUTCONT12020,ReisidazehMokhtariHassaniPedarsaniIEEESP2019}, sometimes also coupled with differential quantization~\cite{KoloskovaStichJaggiICML2019, KoloskovaLinStichJaggiICLR2020, BitsForFree}. However, across these works we find the following design choices: $i)$ non-adaptive systems employing diminishing step-sizes; $ii)$ symmetric combination policies to model the network interactions among agents; $iii)$ strict requirements of local strong convexity, i.e., the cost function of each agent must be strongly convex; and $iv)$ consensus techniques to merge the information shared by the agents across the network.   

In \cite{CarpentieroMattaSayedICASSP2022, CarpentieroMattaSayed2022}, we extended the analysis of distributed learning under communication constraints by removing the aforementioned limitations and considering instead: $(a)$ adaptive systems exploiting stochastic gradient descent with constant step-size which, differently from diminishing step-size implementations, are able to promptly track data drifts and learn continuously; $(b)$ non-symmetric and left-stochastic combination policies to represent a wide variety of real-life network scenarios; $(c)$ strong convexity only at a {\em global level} allowing the existence of convex and non-convex cost functions at a local level; and $(d)$ diffusion (as opposed to consensus) strategies, which have been shown to entail better performance and wider stability range~\cite{Sayed}. We introduced the ACTC (adapt-compress-then-combine) diffusion strategy, which borrows the distributed implementation of the popular ATC (adapt-then-combine) diffusion strategy~\cite{Sayed, ChenSayedTIT2015part1, ChenSayedTIT2015part2} and enriches it to embrace constrained communication. Leveraging a suitable class of {\em randomized differential compression} operators, the ACTC strategy is able to converge to a small neighborhood of the desired solution with significant savings in terms of transmission resources, e.g., quantization bits.

In~\cite{CarpentieroMattaSayed2022, CarpentieroMattaSayedICASSP2022} the learning behavior of the ACTC strategy and its convergence guarantees were characterized in great detail by means of {\em transient} and {\em mean-square stability} analysis. We showed that, despite data compression, {\em it is always possible to achieve mean-square stability by tuning the step-size and the stability parameter of the ACTC algorithm}. We concluded that the peculiar learning behavior of adaptive networks is preserved, exposing two transient phases and a steady-state phase~\cite{ChenSayedTIT2015part1,Sayed,ChenSayedTIT2015part2}. 

After stability and transient analyses, the third essential part that completes the characterization of a learning algorithm is the {\em steady-state performance}. In this work, we fill this research gap, focusing on the relevant case of the so-called {\em MSE (Mean-Square-Error) Networks} \cite{Sayed}. The networks consist of spatially dispersed agents deployed to solve a distributed linear regression problem in an online fashion. This setting represents the distributed counterpart of the popular LMS (least-mean-squares) algorithm employed in signal processing and statistical learning~\cite{sayednewbooks}. One of the earliest works on MSE networks under communication constraints is \cite{ZhaoTuSayedSP2012}, where the compression errors were modeled as noisy sources. In contrast, in this work we start from the design of the compression operators, and take into account the exact form of the associated compression errors. This extension introduces significant additional challenges in the analysis.  

The main contributions of this work are the following.
\begin{itemize}
    \item We provide an upper bound on the mean-square-error of each individual agent for the ACTC strategy over MSE networks. The analysis reveals that the upper bound comprises a term that corresponds to the error achieved in the absence of communication constraints, plus a {\em compression loss}.
    \item By means of the obtained formulas, we establish quantitative relationships between the compression loss and the fundamental attributes of the distributed regression problem, namely, the stochastic approximation error caused by the gradient noise and the network topology (through the Perron eigenvector). 
    \item We show that knowledge of these relations is critical to allocate efficiently the communication resources across the agents, taking into account their individual attributes, such as the quality of their data or their degree of centrality in the network. We propose an optimized allocation strategy, which is fed by an  {\em online} algorithm learning in real time the parameters needed for the optimization. 
    Illustrative examples show significant improvement over a blind (i.e., uniform) resource allocation.
\end{itemize}

\vspace*{10pt}
{\bf Notation}. Boldface letters represent random variables, while normal-font letters represent realizations. Capital letters denote matrices and lowercase letters denote vectors and scalars. This convention is violated in some cases, for example when we denote the number of network agents by $N$. All vectors are intended to be column vectors. The symbol $\mathds{1}_L$ denotes an $L\times 1$ vector with all entries equal to $1$, while the symbol $0$ will be also used to denote all-zeros vectors or matrices. The $L\times L$ identity matrix is denoted by $I_L$. The operation $\textnormal{col}\{x_1,x_2,\ldots,x_L\}$ produces a column vector obtained by stacking the vectors $x_1,x_2,\ldots,x_L$. 
Likewise, the operation $\textnormal{blkdiag}\{X_1, X_2,\ldots, X_N\}$ builds a block-diagonal matrix with matrices $X_1, X_2, \ldots, X_N$ on the main diagonal.
Given two matrices $X$ and $Y$, the notation $X \succeq Y$ indicates that $X - Y$ has nonnegative entries. For square matrices, the notation $X \geq Y$ indicates that $X - Y$ is positive semidefinite. 
The notation $\|x\|$ denotes the $\ell_2$ norm of vector $x$ and $\|X\|$ denotes the corresponding induced norm of matrix $X$. 
Given a vector $x$ and a positive semidefinite real matrix $Y$, the symbol $\|x\|^2_Y$ denotes the weighted norm $x^{\top} Y x$.
The Kronecker product is denoted by $\otimes$ and $X^\top$ denotes the transpose of matrix $X$. The symbol $\E$ denotes the expectation operator. 
Finally, given a a nonnegative function $f(\mu)$, the notation $f(\mu)=O(\mu)$ indicates that there exists a constant $C>0$ and a value $\mu_0$ such that $f(\mu)\leq C \mu$ for all $\mu\leq\mu_0$.

\section{Background} \label{sec:background}
We consider a distributed regression problem tackled by a network of $N$ agents. Agent $k$ at time $i$ observes a {\em response} variable, $\bm{d}_{k,i} \in \mathbb{R}$, and a vector of {\em regressors}, $\bm{u}_{k,i} \in \mathbb{R}^M$, with correlation matrix $R_{u,k} = \E[ \bm{u}_{k,i}\bm{u}_{k,i}^{\top}]$. 
The response and regressors are related through the model:
\beq 
\bm{d}_{k,i} = \bm{u}_{k,i}^{\top}w^o + \bm{v}_{k,i},
\label{eq:linRegModel}
\eeq 
where $\bm{v}_{k,i}$ represents a zero-mean noise process with $\E[ \bm{v}_{k,i}^2] = \sigma^2_{v,k}$ and independent from the regressors. We assume the processes $\{\bm{u}_{k,i}\}$ and $\{\bm{v}_{k,i}\}$ are independent over time and across the agents. 
In \eqref{eq:linRegModel}, $w^o \in \mathbb{R}^M$ is the unknown deterministic parameter vector that is the very object of the cooperative inference process. 
We note in passing that our analysis can be generalized to the case where different agents have different models $w^o_k$, as done, e.g., in~\cite{Sayed, CarpentieroMattaSayedICASSP2022, CarpentieroMattaSayed2022, CarpentieroMattaSayedICASSP2023}. 
We focus here on the important case of a uniform model, since it is more appropriate to examine the case of multiple local models in the multitask framework~\cite{MultiTaskMag}.

A local risk function is associated with every agent in the network, which is defined as the expectation of a quadratic loss function, namely,
\beq
J_k(w)\triangleq \E \left( \bm{d}_{k,i} - \bm{u}_{k,i}^{\top}w\right)^2.
\label{eq:costFunQuad}
\eeq
Through cooperation among agents, the network objective is to minimize the {\em aggregate} cost:
\beq
J(w)=\sum_{k=1}^N p_k J_k(w),
\label{eq:globalCost}
\eeq
where the $\{p_k\}$ are positive coefficients that add up to one. We explain in the sequel that: $i)$ these coefficients are related to the topological properties of the network, leading to useful links between the network arrangement and the distributed optimization problem; and $ii)$ if the regressors of at least one agent have a non-singular correlation matrix, then the unique minimizer of $J(w)$ will coincide with the true model parameter $w^o$.

In our model the agents are arranged in a network, whose topology is described by a {\em combination matrix} $A=[a_{\ell k}]$. The matrix entries $a_{\ell k}$ act as {\em combination weights} placed on the directional edges of the network. When no communication link exists between agents $\ell$ and $k$, the corresponding combination weights $a_{\ell k}$ and $a_{k \ell}$ will be zero. On the other hand, when information can flow only in one direction, say, from $\ell$ to $k$, we will have $a_{\ell k}>0$ and $a_{k\ell}=0$. We will denote the directed neighborhood of agent $k$ (possibly including the self-loop $\ell = k$) by:
\beq
\mathcal{N}_k\triangleq\{\ell \,|\, a_{\ell k}>0\}.
\eeq
We assume the following regularity conditions on the combination matrix.

\begin{assumption}[\bf{Left-stochastic combination matrix}]
\label{Stochastic combination matrix}
The entries on each column of $A$ add up to one. More specifically, it holds that
\begin{equation}\label{combWeights}
a_{\ell k} \geq 0, \quad \sum_{\ell\in\mathcal{N}_k}^N a_{\ell k} = 1,\quad a_{\ell k}=0 \textnormal{ for }\ell\notin\mathcal{N}_k.
\end{equation}~\hfill$\square$
\end{assumption}

\begin{assumption}[{\bf Strongly connected network}]
\label{Strong Connectivity}
A directed path exists linking any node $\ell$ to any other node $k$, which begins at $\ell$ and ends at $k$. 
Moreover, at least one agent $m$ has a self-loop ($a_{mm}>0$).~\hfill$\square$ 
\end{assumption}

When Assumptions~\ref{Stochastic combination matrix} and~\ref{Strong Connectivity} are verified, the combination matrix $A$ turns out to be a primitive matrix~\cite{Sayed,sayednewbooks}. In view of the Perron-Frobenius theorem, $A$ will have a single maximum eigenvalue $1$ corresponding to an eigenvector $\pi=[\pi_1,\pi_2,\ldots,\pi_N]^{\top}$, referred to as the {\em Perron eigenvector}, with all strictly positive entries adding up to one:
\beq
A \pi =\pi,\qquad \pi \succ 0, \qquad \mathds{1}_N^{\top} \, \pi =1.
\label{eq:Perronvecfirstdef}
\eeq

\subsection{The ACTC Diffusion Strategy}
\label{sec:ACTCstrat}
Consider the gradient of $J_k(w)$ in \eqref{eq:costFunQuad}:
\beq 
\nabla J_k(w) = 2\left( R_{u,k}w - r_{du,k} \right),
\label{eq:realGrad}
\eeq 
and introduce the following instantaneous gradient approximation for it at agent $k$ at time $i$:
\beq 
\bm{g}_{k,i}(w) = 2\,\bm{u}_{k,i}\left[\bm{u}_{k,i}^{\top}w - \bm{d}_{k,i} \right].
\label{eq:instGradApprox}
\eeq
In this work we focus on the ACTC strategy proposed in \cite{CarpentieroMattaSayedICASSP2022, CarpentieroMattaSayed2022}, which is described by the following three steps performed iteratively for all time instants $i=1,2,\ldots,$ by each agent $k=1,2,\ldots,N$:
\begin{subequations}\label{ACTC}
\begin{align}
&\bm{\psi}_{k,i} = \bm{w}_{k,i-1} - \mu_k\,\bm{g}_{k,i}(\bm{w}_{k,i-1}) \label{eq:adaStep}\\
&\bm{q}_{\ell,i} = \bm{q}_{\ell,i-1} + \zeta\,\bm{Q}_{\ell}(\bm{\psi}_{\ell,i} - \bm{q}_{\ell,i-1}), \;\; \forall \ell \in \mathcal{N}_k \label{eq:compStep}\\
&\bm{w}_{k,i} = \sum_{\ell \in \mathcal{N}_k} a_{\ell k}\bm{q}_{\ell,i} \label{eq:combStep}
\end{align}
\end{subequations}
The recursion in \eqref{ACTC} is characterized by three time-varying vectors:
an intermediate update $\bm{\psi}_{k,i}$, a differentially-compressed update $\bm{q}_{k,i}$, and the currently estimated minimizer $\bm{w}_{k,i}$.
At time $i=0$, each agent $k$ is initialized with an arbitrary vector $\bm{q}_{k,0}$ (with finite second moment).
Then, agent $k$ receives the initial states $\bm{q}_{\ell,0}$ from its neighbors $\ell\in\mathcal{N}_k$ and computes an initial minimizer $\bm{w}_{k,0} = \sum_{\ell \in \mathcal{N}_k} a_{\ell k}\bm{q}_{\ell,0}$.

For every $i>0$, the agents perform the following steps.
\begin{itemize}
    \item {\em Adaptation step \eqref{eq:adaStep}}: each agent $k$ follows, {\em locally}, the direction $-\bm{g}_{k,i}(\cdot)$ (i.e., the descent direction of its stochastic gradient) weighted by the step-size $\mu_k > 0$ to update the past iterate $\bm{w}_{k,i-1}$. This step corresponds to a {\em self-learning} phase.
    \item {\em Compression step \eqref{eq:compStep}}: each agent $k$ compresses the difference between the intermediate update $\bm{\psi}_{k,i}$ and the previous compressed state $\bm{q}_{k,i-1}$. Compression is performed by applying a suitable randomized operator $\bm{Q}_k(\cdot)$, as detailed in Sec.~\ref{sec:Alistarh}. Each agent $k$ receives from its neighbors $\ell\in\mathcal{N}_k$ the compressed differences $\bm{Q}_{\ell}(\bm{\psi}_{\ell,i}-\bm{q}_{\ell,i-1})$ and recovers the compressed states $\bm{q}_{\ell,i}$, for $\ell\in\mathcal{N}_k$, by adding to the previous state $\bm{q}_{\ell,i-1}$ the received compressed differences\footnote{In order to perform the compression step \eqref{eq:compStep} agent $k$ must possess the variables $\{\bm{q}_{\ell,i-1}\}_{\ell\in\mathcal{N}_k}$. 
    This is possible because at $i=0$ agent $k$ possesses $\{\bm{q}_{\ell,0}\}_{\ell\in\mathcal{N}_k}$ and step \eqref{eq:compStep} depends on $\{\bm{q}_{\ell,i-1}\}_{\ell\in\mathcal{N}_k}$ and the transmitted differences $\{\bm{Q}_{\ell}(\bm{\psi}_{\ell,i} - \bm{q}_{\ell,i-1})\}$. Therefore, the sharing of the initial states $\{\bm{q}_{\ell,0}\}_{\ell\in\mathcal{N}_k}$ is enough for agent $k$ to implement \eqref{eq:compStep} for every time instant $i$ by storing in memory only $\{\bm{q}_{\ell,i}\}_{\ell\in\mathcal{N}_k}$.}. We remark that the compressed difference is scaled by a design parameter $\zeta \in (0,1)$ governing the stability of the ACTC strategy --- see~\cite{CarpentieroMattaSayed2022,CarpentieroMattaSayedICASSP2022}. 
    \item {\em Combination step \eqref{eq:combStep}}: each agent $k$ combines its compressed state with the updated compressed states of its neighbors, performing a {\em social learning} step.
\end{itemize}
For later use, it is convenient to introduce the maximum step-size across agents, along with the {\em scaled} step-sizes:
\beq
\mu\triangleq \max_{k=1,2,\ldots,N} \mu_k, \qquad \alpha_k\dfz\frac{\mu_k}{\mu}.
\label{eq:muScaleDef}
\eeq

\subsection{Compression Operators}
\label{sec:Alistarh}
The implementation of the ACTC strategy in \eqref{ACTC} relies on the compression operator $\bm{Q}_k(\cdot)$. Following~\cite{AlistarhNIPS2017, CarpentieroMattaSayedICASSP2022, CarpentieroMattaSayed2022, BitsForFree, WuHuangHuangZhangICMLR2018, ReisidazehMokhtariHassaniPedarsaniIEEESP2019}, we focus on the following relevant class of {\em randomized} compression operators. 

\begin{assumption}[{\bf Compression operators}]
\label{Compression operator}
A randomized compression operator associates to an input $x$ a stochastic output $\bm{Q}(x)$ whose distribution is governed by a conditional probability measure $\mathbb{P}(\cdot|x)$.
With reference to the compression error:
\beq
\bm{e}(x)\triangleq \bm{Q}(x) - x,
\eeq
the considered class of compression operators fulfills the following two conditions:
\begin{align}
&\E\,\bm{e}(x) = 0\,\,\,\,\,\,\,\,\,\,\,\,\,\,\,\,\,\,\,\,\,\,\,\,\,\,\,\,\,\,\,\,\,\,\,\,\,\,\,\,\,\,\,\,\,\,\,\,\,\,\,\,\,\textnormal{[unbiasedness]}
\label{unbiasedComp}\\
&\E\,\|\bm{e}(x)\|^2 \leq \omega\,\|x\|^2\,\,\,\,\,\,\,\,\,\,\,\,\,\,\,\,\,\,\,\,\,\,\,\,\,\,\,\,\,\,\textnormal{[variance bound]}
\label{boundVarianceComp}
\end{align}
where expectations are relative to $\mathbb{P}(\cdot|x)$.
\hfill$\square$
\end{assumption}

The parameter $\omega$ plays a critical role since it quantifies the amount of compression applied to the input $x$. 
Smaller values of $\omega$ correspond, e.g., to finely quantized data with low distortion, as suggested by \eqref{boundVarianceComp}. On the other hand, large values of $\omega$ correspond to coarsely quantized data with large distortion. Several compression operators belong to the class represented by Assumption~\ref{Compression operator}, and in the forthcoming simulations we will use two popular instances, namely, the randomized quantizer in~\cite{AlistarhNIPS2017} and the randomized sparsifier in~\cite{BitsForFree}.

\begin{example}[{\bf Randomized quantizers} \cite{AlistarhNIPS2017}]
\label{ex:alistarhQuant}
\normalfont
Given an input vector $x \in \mathbb{R}^M$, the randomized quantizer represents its Euclidean norm $\|x\|$ with negligible quantization error (e.g., with machine precision) using $h$ bits, and quantizes each component $x_m$ of $x$ separately. One bit is used to encode the sign of $x_m$ and $r$ bits are used to represent $\chi_m = x_m/\|x\| \in [0,1]$. The interval $[0,1]$ is split into $L$ equal subintervals of size $\theta = 1/L$. Accordingly, given $r$ bits, the number of levels is $L = 2^r - 1$. Each $\chi_m$ is then randomly encoded by choosing one of the two endpoints that enclose it. 
Let $j(\chi_m) = \lfloor \chi_m/\theta \rfloor$ be the index of the lower endpoint, which is equal to $y(\chi_m) = j(\chi_m)\,\theta$. The randomized encoding is based on the rule:
\begin{equation}\label{eq:alistarhRand}
\bm{j}_{\rm{tx}}(\chi_m) = 
\begin{cases}
j(\chi_m) + 1, & \text{with prob.} \;\; \frac{\chi_m - y(\chi_m)}{\theta}\\
j(\chi_m), & \text{otherwise}
\end{cases}.
\end{equation} 
Finally, the $m$-th quantized component is given by:
\beq 
[\bm{Q}(x)]_m = \|x\|\,\textnormal{sign}(x_m)\,\bm{j}_{\rm{tx}}(\chi_m)\,\theta, \quad m=1,2,\ldots,M.  
\eeq 
It is shown in~\cite{AlistarhNIPS2017} that the compression parameter of the randomized quantizer is equal to: 
\beq 
\omega = \min\left\{ \frac{M}{(2^r-1)^2}, \frac{\sqrt{M}}{2^r-1}\right\},
\label{eq:alistarhConst}
\eeq 
and the number of bits needed to encode the vector $x$ is equal to~\cite{AlistarhNIPS2017}:
\beq 
h + M\times(r + 1).
\label{eq:alistarhBits}
\eeq~\hfill$\square$
\end{example}

\begin{example}[{\bf Randomized sparsifiers} \cite{BitsForFree}]
\label{ex:randSparsifier}
\normalfont
Given an input $x \in \mathbb{R}^M$, the randomized sparsifier applies the rule 
\beq
\bm{Q}(x) = \frac{M}{S}\,\bm{F}_S(x),
\eeq
where $\bm{F}_S(x)$ randomly selects $S < M$ components of $x$ and masks the remaining to zero. 

It is shown in~\cite{WangniWangLiuZhangNIPS2018} that the compression parameter of the randomized sparsifier is equal to:
\beq 
\omega = \frac{M}{S} - 1,
\label{eq:randSparsOmega}
\eeq
and the communication savings are quantified by the dimensionality reduction factor $S/M$ as done, e.g., for the lossless case in the compressed sensing or analog compression framework~\cite{VerduAnalog, BolcskeiAnalog}.
Noticing that the sparsified vector can be efficiently represented by encoding the values and positions of the non-masked components only, one can also compute the cost in bits by using $h$ bits for the values and $\log_2(M)$ bits for the positions.~\hfill$\square$
\end{example}

The class of compression operators examined in this work has been shown to model faithfully practical compression schemes, and to be well-tailored to the high-dimensional setting often encountered in learning applications~\cite{AlistarhNIPS2017, KoloskovaStichJaggiICML2019, KoloskovaLinStichJaggiICLR2020, BitsForFree, VerduAnalog, BolcskeiAnalog}. 
For example, in the {\em random sparsifier} paradigm, communication savings are obtained by shrinking the high-dimensional variables into significantly smaller subspaces, and the cost of communication is accordingly measured in terms of the achieved dimensionality reduction. 
Likewise, the vector quantization scheme of Example~\ref{ex:alistarhQuant} uses many bits to represent the norm of the high-dimensional vector, and then distributes a few bits across the individual vector components. 

We hasten to add that the class of compression operators presented here is not exhaustive. Other constructions are possible. 
One useful class proposed in~\cite{Scutari} introduces a second parameter into the variance bound \eqref{boundVarianceComp}.
We remark that the tools used in our work can also be applied to study this other type of compression schemes --- see, e.g.,~\cite{EUSIPCO2022}.

\subsection{Compression Operators in the ACTC strategy}
Given a vector $\bm{x}_{k,i}$ of agent $k$ at time $i$, we denote by $\bm{x}_{0:i}$ a vector collecting all $\bm{x}_{k,i}$ for all agents and all instants up to time $i$.
We also introduce the difference variable to be compressed in step \eqref{eq:compStep} of the ACTC strategy, and the associated error:
\beq  
\bm{\delta}_{k,i}\triangleq\bm{\psi}_{k,i}-\bm{q}_{k,i-1},\qquad 
\bm{e}_{k,i}\triangleq \bm{Q}_k(\bm{\delta}_{k,i}) - \bm{\delta}_{k,i}.
\label{eq:quantnoisemaintextapp}
\eeq 
where agents are allowed to employ different compression operators.
Accordingly, for a given input $x$, each agent $k$ employs a compression operator $\bm{Q}_k(\cdot)$ defined by a probability measure $\mathbb{P}_{k}(\cdot|x)$ satisfying Assumption~\ref{Compression operator} with compression parameter $\omega_k$. 
As is standard in distributed optimization with data compression, we assume that the compression operators are memoryless and independent across agents.
\begin{assumption}[{\bf Conditions on compression operators}]
\label{Compression operator2}
The compression process is independent over time and space, i.e., given $\bm{q}_{0:i-1}$ and $\bm{\psi}_{1:i}$, the compressed vector $\bm{Q}_k(\bm{\delta}_{k,i})$ of agent $k$ depends only on the input $\bm{\delta}_{k,i}$, and compressed vectors are generated as:
\beq
\bm{\delta}_{k,i}\stackrel{\mathbb{P}_k(\cdot |\bm{\delta}_{k,i})}{\longrightarrow} \bm{Q}_k(\bm{\delta}_{k,i}),
\eeq
independently across agents. Then, Assumption~\ref{Compression operator} implies:
\begin{align}
&\E\big[\bm{e}_{k,i}\,|\, \bm{q}_{0:i-1},\bm{\psi}_{0:i}\big]
=
\E\big[\bm{e}_{k,i}\,|\, \bm{\delta}_{k,i}\big]=0,
\label{eq:equantzeromean}\\
&\E\big[\|\bm{e}_{k,i}\|^2 \,|\, \bm{q}_{0:i-1},\bm{\psi}_{0:i}\big]
=
\E\big[\|\bm{e}_{k,i}\|^2 \,|\, \bm{\delta}_{k,i}\big]\leq \omega_k \|\bm{\delta}_{k,i}\|^2,
\label{eq:equantvariancebound}\\
&\E\big[\bm{e}_{\ell,i}\bm{e}_{k,i}^{\top} \,|\, \bm{q}_{0:i-1},\bm{\psi}_{0:i}\big]
=
\E\big[\bm{e}_{\ell,i}\bm{e}_{k,i}^{\top} \,|\, \bm{\delta}_{k,i}\big]=0,\,\forall \ell\neq k.
\label{eq:equantzerocorr}
\end{align}
\hfill$\square$
\end{assumption}

\subsection{Properties of Cost Functions and Gradient Noise}
In this section we state two properties relative to the local risk functions and the gradient noise, which will be critical to establish our results.
\begin{property}[{\bf Smooth risks}]
\label{prop:smoothness}
The local risk function \eqref{eq:costFunQuad} is twice-differentiable and its Hessian matrix has eigenvalues uniformly bounded with respect to $w$, namely,
\beq 
\nabla^2 J_k(w) \leq \eta_k I_M, \qquad \eta_k>0.
\label{eq:lipCond}
\eeq 
The aggregate risk function \eqref{eq:globalCost} is additionally $\nu$-strongly convex, namely,
\beq 
\sum_{k=1}^N \alpha_k \pi_k \nabla^2 J_k(w) \geq \nu I_M,\qquad \nu>0,
\label{eq:strongConvCond}
\eeq 
if, and only if, at least one agent has a non-singular correlation matrix $R_{u,k}$. 
Under \eqref{eq:strongConvCond}, the global risk \eqref{eq:globalCost} has a unique minimizer given by the true parameter $w^o$.
\end{property}

\begin{IEEEproof}
Relation \eqref{eq:lipCond} is immediate since from \eqref{eq:realGrad} the local Hessian matrix is $\nabla^2 J_k(w) = 2 R_{u,k}$. 
On the other hand, in view of the strict positivity of $\alpha_k$ and $\pi_k$, relation \eqref{eq:strongConvCond} is verified if, and only if, at least one local Hessian matrix is non-singular, i.e., positive definite.
\end{IEEEproof}
We further introduce the {\em gradient noise}, which embodies the error introduced by the agents when using \eqref{eq:instGradApprox} in place of the actual gradient in the adaptation step \eqref{eq:adaStep}:
\beq
\bm{n}_{k,i}(\bm{w}_{k,i-1}) \triangleq \bm{g}_{k,i}(\bm{w}_{k,i-1}) - \nabla J_k(\bm{w}_{k,i-1}).
\label{gradNoiseDef}
\eeq
We denote its scaled version by
\beq 
\bm{s}_{k,i} \triangleq \alpha_k\bm{n}_{k,i}(\bm{w}_{k,i-1}).
\label{eq:scaledGradNoiseDef}
\eeq 
The gradient noise acts as a disturbance, which causes a persistent fluctuation of the estimated minimizer of \eqref{eq:globalCost} around its true value~\cite{Sayed, ChenSayedTIT2015part1,ChenSayedTIT2015part2,CarpentieroMattaSayed2022, CarpentieroMattaSayedICASSP2022}. The convergence of the ACTC diffusion strategy was already established under standard regularity conditions on the gradient noise in~\cite{CarpentieroMattaSayed2022}. 

\begin{property}[{\bf Gradient noise process}]\label{prop:gradNoise} 
The gradient noise satisfies the following two properties:
\begin{align}
&\E\big[\bm{s}_{k,i} \,|\, \bm{q}_{0:i-1} , \bm{\psi}_{0:i-1}
\big] = 0,
\label{eq:gradzeromean}\\ 
&\E\big[\bm{s}_{k,i}\,\bm{s}_{\ell,i}^{\top} \,|\, \bm{q}_{0:i-1} , \bm{\psi}_{0:i-1}
\big] = 0, \qquad \forall \ell\neq k.
\label{eq:gradNoiseUncorrelated}
\end{align} 
Furthermore, if we denote the $M \times M$ conditional covariance matrix of the gradient noise associated with agent $k$ by
\beq
\E\left[\bm{s}_{k,i}\bm{s}_{k,i}^{\top} \,|\, \bm{q}_{0:i-1} , \bm{\psi}_{0:i-1} \right]=R_{s,k}(\bm{w}_{k,i-1}),
\label{eq:gradNoiseCovIntro}
\eeq
then it is given by
\begin{align}
R_{s,k}(w) & \triangleq 4\,\alpha^2_k\,\sigma^2_{v,k}\,R_{u,k} \nonumber \\
& + 4\,\alpha^2_k\,\E\left[\bm{U}_{k,i}(w - w^o)(w - w^o)^{\top} \bm{U}_{k,i}^{\top}\right],
\label{eq:gradNoiseSingleAgentCovLimit}
\end{align}
where $\bm{U}_{k,i} = \bm{u}_{k,i}\bm{u}^{\top}_{k,i} - R_{u,k}$.
Moreover, in the steady-state (large $i$), the covariance matrix $R_{s,k}(w)$ evaluated at $\bm{w}_{k,i-1}$ is equivalent to the covariance matrix evaluated at the global minimizer $w^o$, up to an $O(\mu)$ perturbation, formally:
\beq
\limsup_{i\rightarrow\infty}
\E\|R_{s,k}(\bm{w}_{k,i-1}) - R_{s,k}(w^o)\|=O(\mu).
\label{eq:covdifflimsupmain}
\eeq
Relation \eqref{eq:covdifflimsupmain} implies in particular:
\begin{align}
&\textnormal{Tr}[R_{s,k}(w^o)]  - O(\mu)\nonumber\\
&\leq\liminf_{i\rightarrow\infty}\E\|\bm{s}_{k,i}\|^2
\leq 
\limsup_{i\rightarrow\infty}\E\|\bm{s}_{k,i}\|^2 \nonumber\\
&\leq \textnormal{Tr}[R_{s,k}(w^o)]  + O(\mu).
\label{eq:meansquaregradmaintheorem}
\end{align}
\end{property}
\begin{IEEEproof}
See Appendix~\ref{app:gradNoiseProp}.
\end{IEEEproof}

\section{Learning Performance of the ACTC Strategy}
The main result of this work is the characterization of the mean-square-error performance of the ACTC strategy for each agent $k$ in the steady-state regime when $i \rightarrow \infty$. Before stating the theorem that illustrates this result, we need to introduce some key quantities. 
Let
\begin{align}
&\Delta_s\triangleq
\zeta\,\textnormal{Tr}\!\left[\!\left(\sum_{k=1}^N \alpha_k \pi_k R_{u,k}\right)^{-1}\!\!\!\left( \sum_{k=1}^N \alpha_k^2 \,\pi_k^2 \,\sigma^2_{v,k} \,R_{u,k}\right) \!\right],
\label{eq:gradnoisetermnewdef}
\\
&\Delta_{\omega}\triangleq\frac{2\,\zeta}{\nu} \sum_{k=1}^N \alpha_k^2 \, \pi_k^2\, \sigma^2_{v,k} \,\omega_k\,\textnormal{Tr}[R_{u,k}] + \frac{c\,\zeta^2}{2\,\nu},
\label{eq:compnoisetermnewdef}
\end{align}
where $c$ is a constant independent of $\mu$ whose value will depend on the compression parameters $\{\omega_k\}_{k=1}^N$.
Let
\beq
\widetilde{\bm{w}}_{k,i} = \bm{w}_{k,i} - w^o,
\label{eq:centeredwmaintext}
\eeq
and
\beq
\mathscr{E}_k^{\rm{inf}} = \liminf_{i \rightarrow \infty}  \E\|\widetilde{\bm{w}}_{k,i}\|^2,\quad
\mathscr{E}_k^{\rm{sup}} = \limsup_{i \rightarrow \infty}  \E\|\widetilde{\bm{w}}_{k,i}\|^2.
\eeq
\begin{theorem}[{\bf Steady-state performance}]\label{th:ACTCMSD}
Assume at least one agent $m$ has a non-singular correlation matrix $R_{u,m}$. 
Under Assumptions~\ref{Stochastic combination matrix}--\,\ref{Compression operator2}, for sufficiently small values of $\mu$ and $\zeta$ such that the ACTC strategy is mean-square stable,\footnote{The conditions on $\mu$ and $\zeta$ for the mean-square stability of the ACTC strategy are already discussed in~\cite{CarpentieroMattaSayed2022,CarpentieroMattaSayedICASSP2022}.} for all $\ell$ and $k$ (including the case $\ell=k$) we have that:
\beq
\limsup_{i\rightarrow\infty}
\left|
\frac{\E \widetilde{\bm{w}}_{\ell,i}^{\top}\widetilde{\bm{w}}_{k,i}}{\sqrt{\E\|\widetilde{\bm{w}}_{\ell,i}\|^2\E\|\widetilde{\bm{w}}_{k,i}\|^2}}
-
1
\right|=O(\mu^{1/2}).
\label{eq:smallCorrRes}
\eeq
Furthermore, the mean-square-error of each agent $k$ is bounded as follows:
\beq
\mu\,\Delta_s - O(\mu^{3/2})
\!\leq
\mathscr{E}_k^{\rm{inf}}
\!\leq
\mathscr{E}_k^{\rm{sup}}
\!\leq 
\mu
\left(
\Delta_s \!+\! \Delta_{\omega}\right)
+ O(\mu^{3/2}).
\label{eq:covRecursTraceSubsTh}
\eeq
\end{theorem}
\begin{IEEEproof}
See Appendix~\ref{app:MSDthProof} at the end of this manuscript, which in turn benefits from general results in Appendix~\ref{app:firstmainapp}.
\end{IEEEproof}

\subsection{Comments on Theorem~\ref{th:ACTCMSD}}
Relation \eqref{eq:smallCorrRes} highlights one fundamental result.
It reveals that, in steady-state and in the considered small step-size regime, the iterates of any two agents tend to be perfectly correlated, i.e.,
\beq
\E \widetilde{\bm{w}}_{\ell,i}^{\top}\widetilde{\bm{w}}_{k,i}
\approx 
\sqrt{\E\|\widetilde{\bm{w}}_{\ell,i}\|^2\E\|\widetilde{\bm{w}}_{k,i}\|^2},
\eeq
implying, in view of the Cauchy-Schwarz (in)equality, that $\widetilde{\bm{w}}_{\ell,i}$ and $\widetilde{\bm{w}}_{k,i}$ are approximately proportional. 
On the other hand, since \eqref{eq:smallCorrRes} holds also for $\ell=k$, we conclude that $\widetilde{\bm{w}}_{\ell,i}$ and $\widetilde{\bm{w}}_{k,i}$ have approximately the same norm and, hence, the proportionality constant is $1$:
\beq
\widetilde{\bm{w}}_{\ell,i} \approx \widetilde{\bm{w}}_{k,i}.
\eeq
This means that the errors of all agents evolve in a {\em coordinated} manner. 

Once established that all agents attain the same error, we want to evaluate this error. 
First, note that the lower bound in \eqref{eq:covRecursTraceSubsTh} is determined by the parameter $\Delta_s$, which from \eqref{eq:gradnoisetermnewdef} is ruled by the main source of variability of the regression problem, namely, the gradient noise. In fact, the lower bound coincides with the limiting MSE obtained for the uncompressed strategy~\cite{Sayed}. In other words, the compressed strategy cannot be better than the uncompressed one. This appears to be an obvious result, which nevertheless requires a formal proof. The upper bound captures the effect of compression through the term $\Delta_{\omega}$ in \eqref{eq:compnoisetermnewdef}.
Observe that the dominant terms in the lower and upper bounds do not coincide. This gap arises since, from Assumption~\ref{Compression operator}, we only have an upper bound characterizing the compression error. 
If a more precise characterization of the error were available, we would be able to fill the gap between the lower and upper bound. 
In the following, we will use the upper bound as a proxy for the MSE performance, since it takes into account the effect of compression, and can therefore be useful also to optimize the allocation of communication resources.

Let us examine more closely the structure of the upper bound in \eqref{eq:covRecursTraceSubsTh}. Exploiting \eqref{eq:gradnoisetermnewdef} and \eqref{eq:compnoisetermnewdef}, it can be conveniently decomposed as follows (neglecting the higher-order $O(\mu^{3/2})$ correction):
\begin{align}
&\underbrace{\mu\,\zeta \,\textnormal{Tr}\left[ \left( \sum_{k=1}^N \alpha_k \pi_k R_{u,k} \right)^{-1} \left( \sum_{k=1}^N \alpha_k^2 \,\pi_k^2 \,\sigma^2_{v,k} \,R_{u,k}\right) \right] }_{\textnormal{uncompressed evolution error}} \nonumber \\ 
& + \underbrace{\frac{2\,\mu\,\zeta}{\nu}\sum_{k=1}^N \alpha_k^2 \,\pi_k^2 \,\sigma^2_{v,k} \, \omega_k\textnormal{Tr}\left[R_{u,k}\right] }_{\textnormal{gradient noise compression loss}} 
 + \underbrace{\frac{c\,\mu\,\zeta^2}{2\,\nu}}_{\substack{\textnormal{network error} \\ \textnormal{compression loss}}}\!\!\!\!\!\!.
\label{eq:theExplan}
\end{align}
We focus on the individual components of the error.

{\em --- Uncompressed evolution error}. This term corresponds to the mean-square-error achieved by the diffusion strategy if perfect (i.e., uncompressed) information is shared. It has the same structure of the classical ATC diffusion strategy performance, involving the main attributes of the inference problem, namely, the regressors' correlation matrices $R_{u,k}$ and the noise variances $\sigma^2_{v,k}$, suitably weighted by the Perron eigenvector and the parameters $\alpha_k$ that quantify the differences among the individual step-sizes~\cite{Sayed, ChenSayedTIT2015part2}.

{\em --- Compression loss}. An additional source of error affects the steady-state behavior because of the sharing of compressed information. 
Equation \eqref{eq:theExplan} highlights the fundamental sources from which the compression loss originates, which are the \emph{gradient noise} and the \emph{network error component}. 

The former term can be related to the behavior of classical quantization systems, whose distortion depends on the variance of the random variable to be compressed. 
In the steady-state regime, the innovation in the compression step \eqref{eq:compStep} has variance related to the gradient noise --- see Lemma~\ref{lem:deltaApprox} in Appendix~\ref{sec:quantnoiseanalysislemma2main}. Accordingly, we find a {\em distortion} term containing the trace of the gradient noise covariance matrix, $R_{s,k}(w^o)=4\,\alpha^2_k\,\sigma^2_{v,k} R_{u,k}$.  
The network error component term deals with the local discrepancies between individual agents and the coordinated evolution of the network towards the common minimizer. In the classical ATC diffusion strategy this error term is a higher-order correction~\cite{Sayed, ChenSayedTIT2015part1, ChenSayedTIT2015part2}, while in the ACTC diffusion strategy it is increased due to the compression error that seeps into the algorithm evolution; it is now on the order of $\mu$ as the other terms in the MSE expression. 

Remarkably, we see from \eqref{eq:theExplan} that the compression parameters $\omega_k$ are weighted by the squared entries of the Perron eigenvector, suggesting a useful relationship between the network arrangement and the design of the compression operators. 
Therefore, designing the compression operators considering the network structure can be used to tune the mean-square-error performance. We will explore this possibility in the next section.

\section{Illustrative Examples and Optimized Resource Allocation}
\label{sec:experiments}
\begin{figure*}[t]
\centering
\includegraphics[width = 0.4\linewidth]{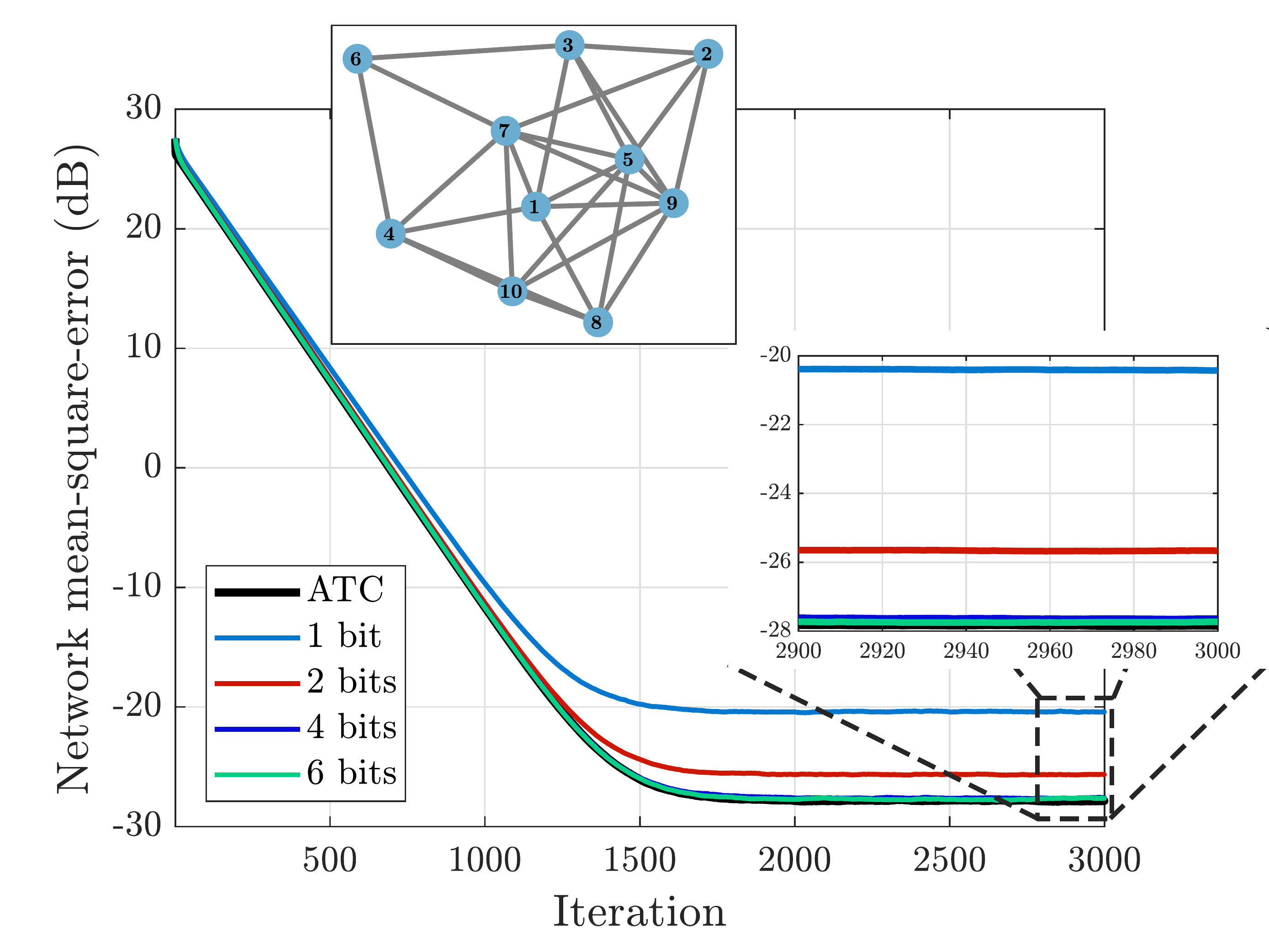}
\includegraphics[width = 0.4\linewidth]{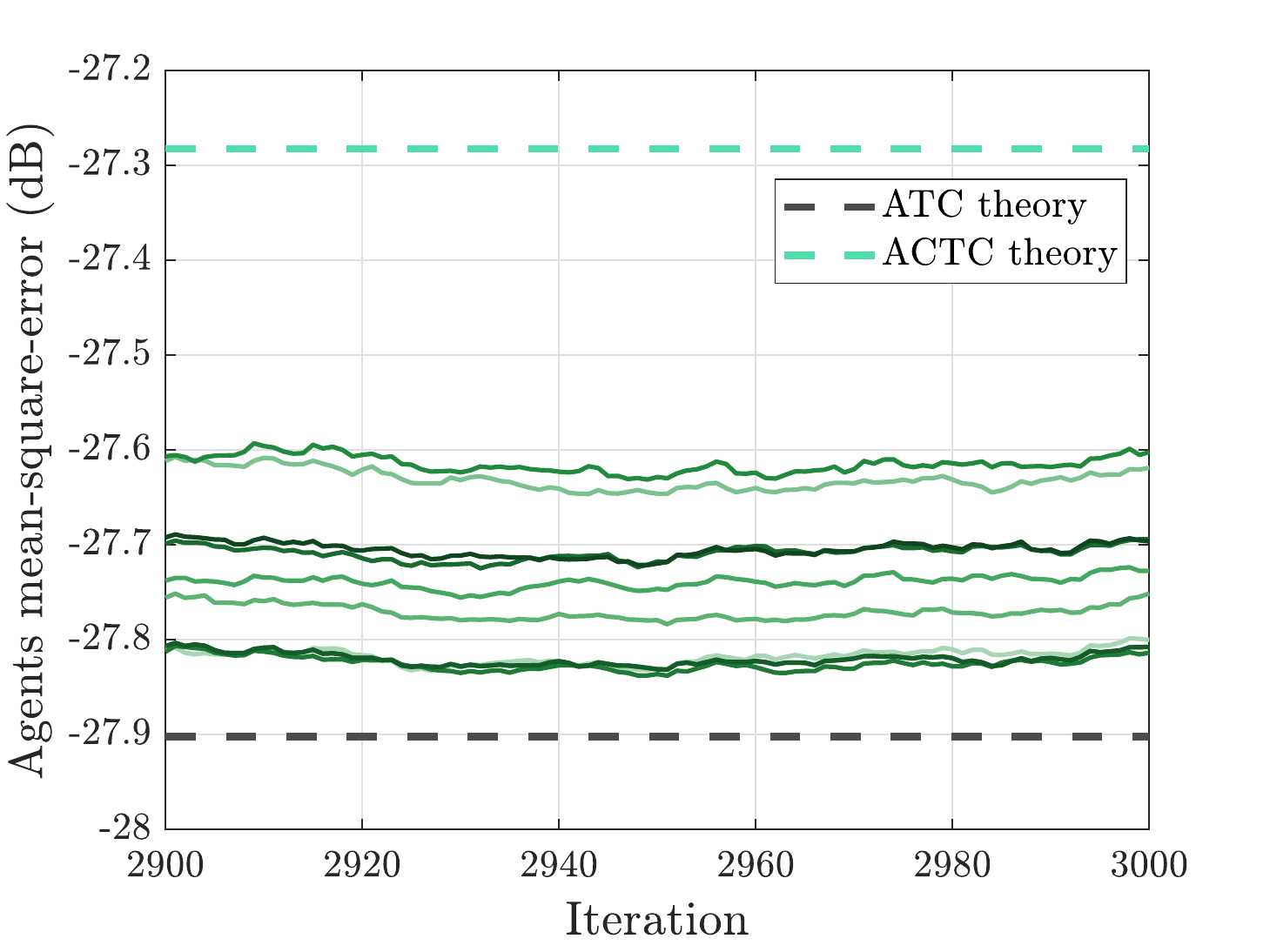}
\caption{ 
ACTC mean-square-error performance, as a function of the iteration $i$. We refer to the experimental setting in Sec.~\ref{sec:experiments}, where: $(i)$ the regressors $\bm{u}_{k,i} \in \mathbb{R}^{M}$ are zero-mean Gaussian with diagonal covariance matrices and variances drawn as independent realizations from a uniform distribution in $(1,4)$; $(ii)$ the noises $\bm{v}_{k,i} \in \mathbb{R}$ are zero-mean Gaussian with variances drawn as independent realizations from a uniform distribution in $(0.25,1)$. The ACTC strategy is run with equal step-sizes $\mu_k=\mu= 10^{-2}$ and stability parameter $\zeta = 10^{-1}$. 
{\em Left plot}. Network mean-square-error when agents apply the randomized quantizers from Example~\ref{ex:alistarhQuant} with different bit-rates. The inset plot shows the network topology, on top of which the averaging combination policy is applied. All nodes have a self-loop (not shown, for simplicity). 
{\em Right plot}. Zoom on the steady-state  mean-square-error of the agents (shown in different shades of green) when employing $r=6$ bits. The black dashed line represents the mean-square-error of the ATC strategy~\cite{Sayed, ChenSayedTIT2015part2} and the green dashed line represents the mean-square-error bound in \eqref{eq:covRecursTraceSubsTh}. The mean-square-error is estimated by means of $10^3$ Monte Carlo runs.
}
\label{fig:1}
\end{figure*}

We will illustrate the implications of Theorem~\ref{th:ACTCMSD} in the following MSE network scenario: the observed regressors have dimensionality $M=30$ and are characterized by diagonal matrices $R_{u,k}$ with variances uniformly and independently drawn in the interval $(1,4)$. Likewise, the noise variances $\sigma_{v,k}^2$ are drawn uniformly in the interval $(0.25,1)$. The network is composed of $N=10$ agents, arranged according to the topology depicted in Fig.~\ref{fig:1}, on top of which we set a left-stochastic combination policy using the uniform averaging rule~\cite{Sayed}. All agents employ the same step-size $\mu_k = \mu = 10^{-2}$ and the stability parameter is $\zeta =  10^{-1}$. 
The left plot in Fig.~\ref{fig:1} shows how data compression affects the \emph{network} performance, i.e., the MSE averaged over all the agents:
\beq 
\frac{1}{N}\sum_{k=1}^N\E\|\bm{w}_{k,i} - w^o\|^2.
\eeq 
The agents quantize the transmitted information using the randomized quantizers presented in Example~\ref{ex:alistarhQuant}. The ACTC performance approaches the ATC performance as the bit-rate increases~\cite{CarpentieroMattaSayed2022, CarpentieroMattaSayedICASSP2022}, in accordance with \eqref{eq:covRecursTraceSubsTh}. 

\subsection{Optimized Resource Allocation}
\label{sec:resAlloc}
Given a total budget of communication resources, we wish to optimize their allocation across agents in view of the learning performance. Let $x = [x_1, x_2, \ldots, x_N]$, where $x_k$ represents the communication resources assigned to agent $k$. Note that the role of the variables $x_k$ depends on the specific compression operator chosen. For instance, when using the randomized quantizers in Example~\ref{ex:alistarhQuant}, $x_k$ is the number of bits to quantize the scaled components of the input. 
When using the randomized sparsifiers in Example~\ref{ex:randSparsifier}, $x_k$ represents the number of components of the input vector that are not masked to zero by agent $k$.
We assume that each $x_k$ must lie in a given interval, and the total budget must not exceed a prescribed value $X$. 
Under these constraints we can minimize the upper bound in \eqref{eq:covRecursTraceSubsTh}. 
Removing from \eqref{eq:covRecursTraceSubsTh} the term $\Delta_s$ that is independent of $x$, and neglecting the $O(\mu^{3/2})$ correction and the network error component compression loss term $c\,\zeta/(2\nu)$ whose impact on the performance is significantly reduced for the considered small values of $\zeta$, we obtain the problem:
\beq
\begin{aligned}
\min_{x \in \mathbb{R}^N} & \sum_{k=1}^N \pi_k^2\,\omega_k \,d_k\nonumber\\
\textnormal{s.t.} \quad & \sum_{k=1}^N x_k \leq X\\
  &  x_{\rm{min}}  \leq x_k \leq x_{\rm{max}}, \; k=1,2,\ldots,N.   
\end{aligned}
\label{eq:secondOptProblem}
\eeq
where we recall that $\omega_k$ depends on $x_k$ and where we set:
\beq
d_k\triangleq \frac {\textnormal{Tr}[R_{s,k}(w^o)]}{4}=\alpha_k^2\, \sigma^2_{v,k} \,\textnormal{Tr}[R_{u,k}].
\label{eq:dkdistortiondef}
\eeq 
The exact solution of \eqref{eq:secondOptProblem} would require to enforce an integer constraint on $x$, leading to nonpractical solvers.
As is typical in bit-allocation problems pursued in the theory of quantization~\cite{GershoGrayBook}, we solve problem \eqref{eq:secondOptProblem} over the real domain and then round the solution to integer values that satisfy the constraints. 

\subsection{Online Resource Allocation Solution}
We can compute the solution to \eqref{eq:secondOptProblem} by applying the Karush-Kuhn-Tucker (KKT) conditions, see Appendix~\ref{app:KKT}.
The solution requires knowledge of the Perron eigenvector entries $\pi_k$ and of the distortion coefficients $d_k$, which are in general not available to agents, but can be estimated in an online manner, as we now show. 

One way to compute an estimate of the Perron eigenvector is by running, alongside the ACTC diffusion strategy, an averaging consensus algorithm~\cite{XiaoBoydSCL2004}. 
After $t$ iterations, the output of the consensus algorithm can be cast in the form:
\beq 
y_t = (A^{\top})^t \,y_0\overset{t\rightarrow\infty}{\longrightarrow} \sum_{\ell=1}^N \pi_{\ell} y_{\ell,0},
\label{eq:consensusIter}
\eeq 
where $y_t=[y_{k,t}]$ is the $N\times 1$ vector collecting the output $y_{k,t}$ of agent $k$ at iteration $t$. 
Under Assumption~\ref{Stochastic combination matrix}, the columns of the powers of the combination matrix $A$ converge to the Perron eigenvector~\cite{Sayed}. 
If agent $k$ is initialized with value $y_{k,0}=1$ and all other agents with value $0$, all agents will converge to the $k$-the entry of the Perron eigenvector, $\pi_k$.
Repeating the process $N$ times yields the estimate of $\pi$ at all agents.
Note that averaging consensus converges exponentially fast~\cite{XiaoBoydSCL2004}, allowing to obtain a faithful estimate of $\pi$ in a few iterations. 

To estimate the distortion terms $d_k$ we follow \cite{Sayed, ZhaoTuSayedSP2012}. 
First, we observe that from \eqref{eq:adaStep} we can write
\begin{align}
&\bm{\psi}_{k,i} - \bm{w}_{k,i-1} = \mu_k \,\bm{g}_{k,i}(\bm{w}_{k,i-1}) \nonumber \\
&=\mu\,\bm{s}_{k,i}+ \mu\,\alpha_k\big(\nabla J_k(\bm{w}_{k,i-1}) - \nabla J_k(w^o)\big),
\label{eq:usefuldiffonline}
\end{align}
where in the second equality we add and subtract the gradient $\nabla J_k(\bm{w}_{k,i-1})$, and use the fact that $\nabla J_k(w^o)=0$.
From the Lipschitz condition in Property~\ref{prop:smoothness}, we can bound $\|\nabla J_k(\bm{w}_{k,i-1}) - \nabla J_k(w^o)\|^2$ in terms of the norm $\|\bm{w}_{k,i-1} - w^o\|^2$, whose limit superior is $O(\mu)$. Therefore, using  \eqref{eq:meansquaregradmaintheorem} and \eqref{eq:dkdistortiondef}, from \eqref{eq:gradzeromean} and \eqref{eq:usefuldiffonline} we obtain:
\begin{align}
&d_k - O(\mu)\leq\frac{4}{\mu^2}\liminf_{i\rightarrow\infty}\E\|\bm{\psi}_{k,i} - \bm{w}_{k,i-1}\|^2
\nonumber\\
&\leq 
\frac{4}{\mu^2}\limsup_{i\rightarrow\infty}\E\|\bm{\psi}_{k,i} - \bm{w}_{k,i-1}\|^2 
\leq d_k  + O(\mu).
\label{eq:miracolousApprox}
\end{align}
Accordingly, in steady-state (i.e., for $i$ sufficiently large) and in the considered regime of small step-sizes we have approximately:
\beq 
\E\|\bm{\psi}_{k,i} - \bm{w}_{k,i-1}\|^2 \propto d_k 
\label{eq:miracolousApprox}
\eeq 
The proportionality constant is clearly immaterial for the optimization in \eqref{eq:secondOptProblem}.
Under ergodicity, the statistical mean in \eqref{eq:miracolousApprox} can be estimated by evaluating the (online) empirical mean of $\|\bm{\psi}_{k,i} - \bm{w}_{k,i-1}\|^2$ as time progresses, or even, in an adaptive perspective, resorting to a smoothing filter with forgetting factor $\xi_k \in (0,1)$, namely\cite{Sayed, ZhaoTuSayedSP2012}:
\beq 
\bm{d}_{k,i} = (1 - \xi_k)\,\bm{d}_{k,i-1} + \xi_k \|\bm{\psi}_{k,i} - \bm{w}_{k,i-1}\|^2,
\label{eq:dkEstimator}
\eeq 
where $\bm{d}_{k,i}$ is the estimate of $d_k$ at time $i$. 
In order to apply the KKT solution provided in Appendix~\ref{app:KKT} each agent must have access to the estimated $\{\pi_k\}$ and $\{d_k\}$. Accordingly, once the agents have estimated these quantities, they share these values over the network. 
In terms of communication cost: $i)$ the estimate of $d_k$ is performed locally (no communication) by each agent; $ii)$ the communication cost associated with the consensus algorithm used to estimate the Perron eigenvector, as well as to sharing the estimated quantities, is safely neglected since it takes place only once and is therefore washed out over the iterations of the ACTC strategy.

\begin{figure}[t]
\centering
\includegraphics[width=0.5\columnwidth]{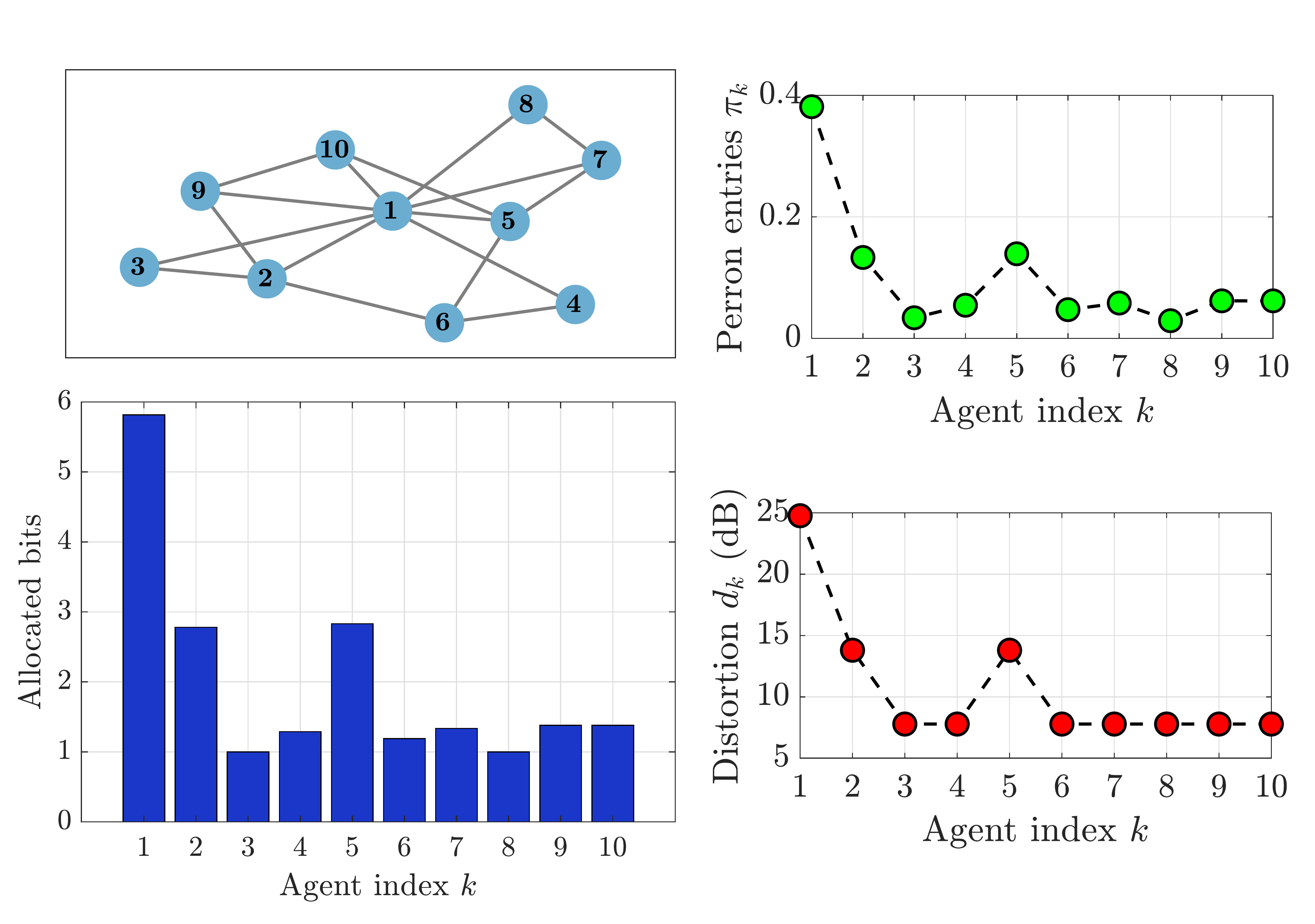}
\caption{
{\em Top left.} Network topology built using the Bollob\'{a}s-Riordan model, where agent $1$ acts as a hub node. 
{\em Bottom left.} Optimized bit allocation for the ACTC strategy using the randomized quantizers in Example~\ref{ex:alistarhQuant}, with the solution to \eqref{eq:secondOptProblem} from Appendix~\ref{app:KKT}. 
{\em Top right}.
Distribution of the Perron weights $\pi_k$ for the considered topology equipped with a relative degree combination policy. 
{\em Bottom right}. Distortion values $d_k$ from \eqref{eq:dkdistortiondef}. 
We see how the bits assigned to each agent follow the trend of the coefficients $\pi_k$ and $d_k$. In this example we set the matrices $R_{u,k}$ as $\{5I_M, 2I_M, I_M, I_M, 2I_M, I_M, I_M, I_M, I_M, I_M\}$ and the variances $\sigma^2_{v,k}$ as $\{1, 0.2, 0.1, 0.1, 0.2, 0.1, 0.1, 0.1, 0.1, 0.1\}$.}
\label{fig:2}
\end{figure}

\subsection{Application to Practical Compression Schemes}
We first examine the solution to \eqref{eq:secondOptProblem} when employing the randomized quantizers in Example~\ref{ex:alistarhQuant} and the randomized sparsifiers in Example~\ref{ex:randSparsifier}. In the following examples we generate the network topology, shown in Fig.~\ref{fig:2}, using the Bollob\'{a}s-Riordan preferential-attachment model~\cite{BollobasRiordanSpencerTusnady2001, BollobasRiordan2003} which allows us to create a naturally clustered topology with {\em hub} nodes having large neighborhoods. On top of the network topology we set a combination policy obtained through the relative degree policy \cite{Sayed}. 

\subsubsection{Resource Allocation with Randomized Quantizer}
Substituting expression \eqref{eq:alistarhConst} into \eqref{eq:secondOptProblem} leads to a nonpractical non-differentiable and non-convex optimization problem. Therefore, we consider the approximate version of the compression parameter \eqref{eq:alistarhConst} given by:
\beq 
\widehat{\omega}_k = \frac{M}{(2^{x_k}-1)^2}.
\label{eq:approxCompAlistarh}
\eeq
According to \eqref{eq:alistarhConst}, relation \eqref{eq:approxCompAlistarh} is an upper bound on the compression error \eqref{boundVarianceComp}, which is known to be tight in  the {\em high resolution} regime, i.e., for relatively high values of $x_k$. 
We compute the solution to problem \eqref{eq:secondOptProblem} by applying the Karush-Kuhn-Tucker (KKT) conditions --- see Appendix~\ref{app:KKT}.

Under the high resolution approximation implied by \eqref{eq:approxCompAlistarh}, problem \eqref{eq:secondOptProblem} can be solved in closed form by resorting to classical methods for optimal bit allocation with scalar quantizers. 
The closed-form solution will give us an insightful interpretation of the optimized allocation. Writing \eqref{eq:approxCompAlistarh} as $\widehat{\omega}_k \approx M 2^{-2x_k}$ and neglecting the box constraints, problem \eqref{eq:secondOptProblem} has the same structure as the bit allocation problem in~\cite{GershoGrayBook}. By means of Lagrange multipliers, or applying the arithmetic/geometric mean inequality, the optimal allocation can be shown to be~\cite[\S 8.3]{GershoGrayBook}:
\beq 
x_k = \bar{x} + \log_2{\frac{\pi_k}{\pi_{\rm{av}}}} + \frac{1}{2}\log_2{\frac{d_k}{d_{\rm{av}}}},
\label{eq:gershoGrayAllocation}
\eeq 
where
\beq
\bar{x} = \frac{X}{N}, \quad \pi_{\rm{av}} = \left( \prod_{k=1}^N \pi_k \right)^{\frac 1 N}, \quad d_{\rm{av}} = \left( \prod_{k=1}^N d_k \right)^{\frac 1 N}.
\label{eq:geomMeanPerron}
\eeq
Recognizing that $\bar{x}$ is the average bit allocation per agent, this result shows that the resources available to agent $k$ are a perturbation of $\bar{x}$ depending on the Perron weights $\pi_k$ and the distortion values $d_k$.

Observing the optimized bit allocation for problem \eqref{eq:secondOptProblem} shown in the example of Fig.~\ref{fig:2}, we notice that most of the available resources are assigned to agents with high Perron weights (i.e., with high centrality) and significant error sources (i.e., more error-prone). 
This interpretation agrees with relation \eqref{eq:gershoGrayAllocation}, which reveals precisely how network centrality and data quality interplay to determine the optimized resource allocation. 
The higher the centrality of agent $k$ is, the greater $\pi_k$ will be, implying additional compression resources with respect to the average allocation $\bar{x}$ when $\pi_k > \pi_{\rm{av}}$. 
Indeed, ``central'' agents forward information to many neighbors, being very influential in driving the network toward the desired minimizer. This explains why in \eqref{eq:gershoGrayAllocation} a higher Perron weight $\pi_k$ favors the assignment of more bits. 
By the same token, it is beneficial to assign more communication resources to agents characterized by a distortion $d_k$ higher than the average distortion $d_{\rm{av}}$, since in this manner the compression errors that propagate across the network are reduced. 
In the presence of conflicting requirements (e.g., low centrality and high distortion), the allocation rule \eqref{eq:gershoGrayAllocation} reveals how to manage the trade-off.

\begin{figure*}[t]
\centering
\includegraphics[width = 0.4\linewidth]{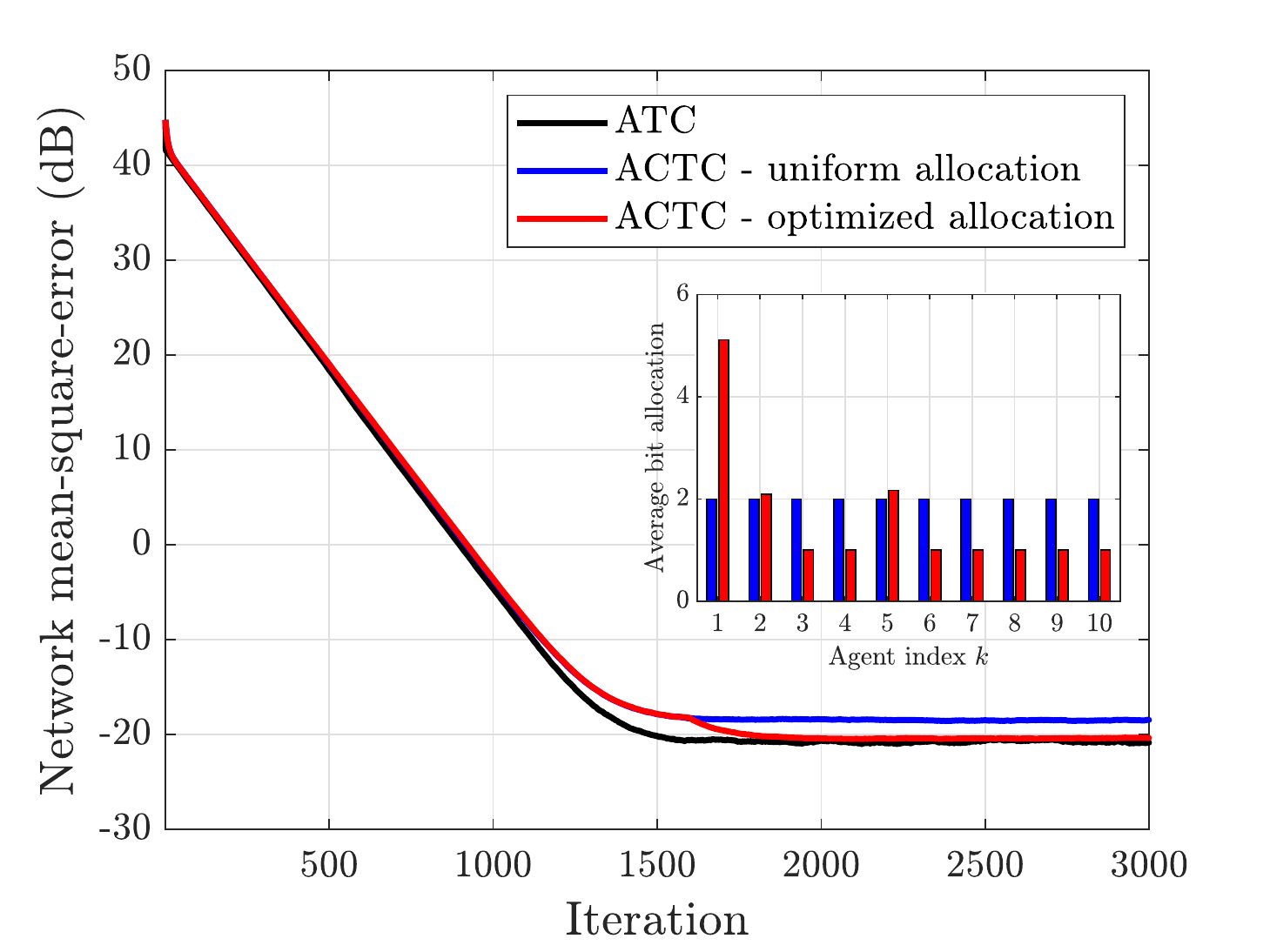}
\includegraphics[width = 0.4\linewidth]{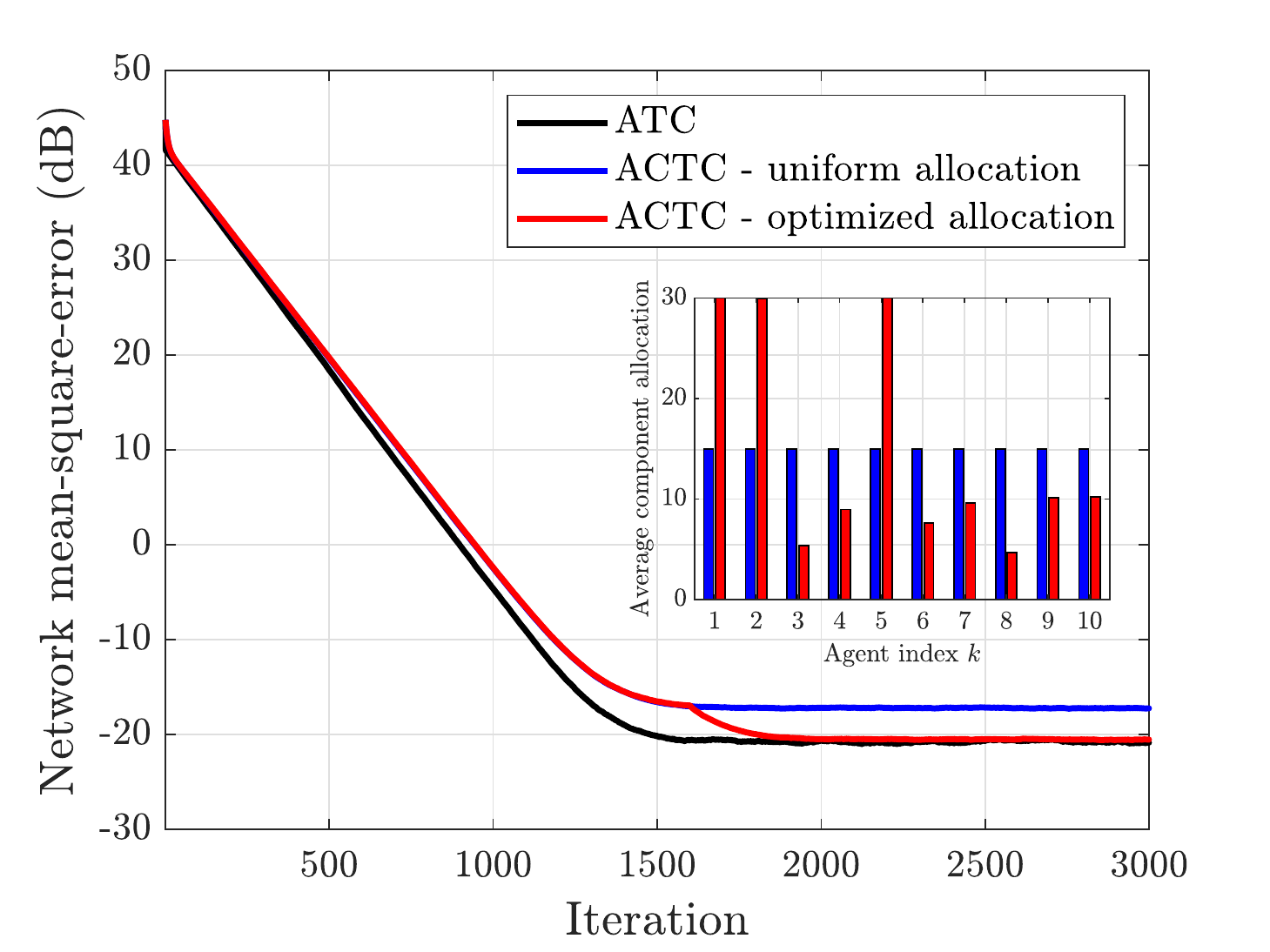}
\caption{ 
ACTC mean-square-error performance, as a function of the iteration $i$, with uniform and optimized resource allocation. The network topology is shown in Fig.~\ref{fig:2}, and is equipped with a relative degree combination policy. The regressors' matrices $R_{u,k}$ and the noise variances $\sigma^2_{v,k}$ are also reported in the caption of Fig.~\ref{fig:2}. The other system parameters are the same used in the example of Fig.~\ref{fig:1}.
{\em Left plot}. Network mean-square-error when agents apply the randomized quantizers in Example~\ref{ex:alistarhQuant}, with a total resource budget $X=20$ and with constraints $1 \leq x_k \leq 11$. The inset plot shows the bit-rates $x_k$ for both allocation strategies. {\em Right plot}. Network mean-square-error when agents apply the randomized sparsifiers in Example~\ref{ex:randSparsifier} with a total resource budget $X=150$ with constraints $1 \leq x_k \leq M$. The inset plot shows the number of non-masked components $x_k$ for both allocation strategies. In both experiments, agents compute the optimized resource allocation at time $T_{\rm{opt}}=1600$. The mean-square-error is estimated by means of $10^3$ Monte Carlo runs.
}
\label{fig:3}
\end{figure*}

Figure~\ref{fig:3} shows that the optimized resource allocation produces a significant improvement of the steady-state network performance of the ACTC strategy, which moves closer to the reference ATC performance {\em without any additional resource expense}. The left panel shows the effect of the optimized allocation for the randomized quantizer, setting in \eqref{eq:secondOptProblem} $X = 20$, $x_{\rm{min}} = 1$ and $x_{\rm{max}} = 11$. 
Agents solve the optimization problem at time $T_{\rm{opt}}=1600$, after entering the steady-state. 
The blue and red curves compare, respectively, the uniform allocation, $x_k = \bar{x}$ for all $k$, against the optimized allocation obtained by finding the KKT solution and then applying the floor operator to obtain integer values.
We see how the network benefits from the optimized allocation. Notably, the red curve approaches the reference ATC error (i.e., the error of the uncompressed strategy) in a few steps after time instant $T_{\rm{opt}}$.
The optimized policy reduces the mean-square-error by approximately $2$ dB with respect to the performance achieved with uniform allocation. 

It is also useful to compare the bit expense of the ACTC and ATC strategies. 
Assuming a machine precision $h = 32$ and considering $T = 2000$ iterations to reach convergence, from \eqref{eq:alistarhBits} we see that the bit expense of the ACTC strategy is:
\beq
r_{\rm{ACTC}} = T \left(
N h + M X + M N \right)
= 2.396 \;\textnormal{Mbits},
\label{eq:ACTCBitsCost}
\eeq
Note that the bit expense depends only on the total bit budget $X$ and, hence, is the same for the uniform and optimized allocation. However, the fundamental difference is that, with the same expense, the optimized allocation is able to attain almost the same performance as reference (uncompressed) ATC strategy, whose bit expense is:
\beq
r_{\rm{ATC}}  = T \times N \times M \times h = 19.2 \;\textnormal{Mbits},
\label{eq:ATCBitsCost}
\eeq
implying a significant saving of more than $16\;\textnormal{Mbits}$.

\subsubsection{Resource Allocation with Randomized Sparsifiers}
We find the solution to problem \eqref{eq:secondOptProblem} when using the compression parameter \eqref{eq:randSparsOmega} of the randomized sparsifier applying again the KKT conditions --- see Appendix~\ref{app:KKT}. We set a total budget of resources $X = 150$ with $x_{\rm{min}} = 1$ and $x_{\rm{max}} = M$. As in the previous example, the optimized resource allocation is computed at $i=1600$. In the uniform allocation each agent is equipped with $x_k = 150/10=15$ components of the input vector that are non-masked to zero. The right panel of Fig.~\ref{fig:3} compares the performance of the ACTC strategy that employs the randomized sparsifier with uniform allocation against the KKT optimized solution. Once again the optimized policy improves the learning performance, now reducing the mean-square-error by approximately $3$ dB with respect to the performance achieved with uniform allocation.
Note that this result is achieved by guaranteeing a dimensionality reduction factor $X/M=1/2$, as compared to the reference ATC strategy which transmits all vector components for all agents. 

\section{Conclusion}
This work focused on an online distributed regression problem under communication constraints. 
We have characterized the performance of the adapt-compress-then-combine (ACTC) diffusion strategy by establishing an upper bound on the MSE of each individual agent. 
The analysis reveals that all agents learn in a coordinated manner by attaining the same steady-state MSE, which is composed by a term characterizing the classical (i.e., no data compression) ATC diffusion strategy, plus an additional compression loss. 
The obtained analytical solution allows us to characterize the compression loss in terms of the network topology and the gradient noise, and enables the possibility of optimizing the bit allocation across the agents, depending on their data quality and centrality in the network. 
To this end, we have devised an optimization procedure alongside an online strategy to learn the parameters necessary for the optimization procedure. The experimental analysis showed significant savings in terms of communication resources. 
There is room for several useful extensions, including: the characterization of the MSE performance for other cost functions; the extension to other models for the compression operators and variable-rate quantization schemes; the extension to learning problems under subspace constraints and multitask settings.

\appendices

\section{Useful Recursions}
\label{app:firstmainapp}
In the following treatment, given a collection of $M\times 1$ vectors $\bm{x}_{k,i}$, we will often work in terms of extended $MN\times 1$ vectors:
\beq
\bm{x}_i \triangleq \textnormal{col}\{\bm{x}_{1,i}, \bm{x}_{2,i}, \ldots, \bm{x}_{N,i}\}.
\eeq
In order to evaluate the mean-square error of the iterates $\bm{w}_{k,i}$, it is convenient to focus first on the mean-square-error of the {\em compressed} iterates $\bm{q}_{k,i}$, and in particular on the centered vectors $\widetilde{\bm{q}}_{k,i}=\bm{q}_{k,i} - w^o$, whose corresponding $MN\times 1$ extended vector will be denoted by $\widetilde{\bm{q}}_i$.
From the ACTC recursion \eqref{ACTC}, it is straightforward to verify that $\widetilde{\bm{q}}_i$ satisfies the following recursion (the details of the calculations can be found in Sec.~\ref{app:collectionA}):
\begin{align}
\widetilde{\bm{q}}_i & = \mathcal{B}\,\widetilde{\bm{q}}_{i-1} + \zeta\,\bm{e}_i - \mu\,\zeta\,\bm{s}_{i},
\label{eq:qTildeTotalMain}
\end{align}
where the $MN\times MN$ matrix $\mathcal{B}$ is defined in \eqref{eq:qTildeTotal}.
In \eqref{eq:netACTC1} we also obtain the following recursion for the network-level quantity $\bm{\delta}_i$
\beq
\bm{\delta}_{i} = [(I_{MN} - \mu\,\mathcal{H})\,\mathcal{A}^{\top}  -  I_{MN}]\widetilde{\bm{q}}_{i-1} 
- \mu\,\bm{s}_{i},
\label{eq:deltarecmain33}
\eeq
where the matrices $\mathcal{H}$ and $\mathcal{A}$ are defined in \eqref{hCal}.

It is also useful to introduce the {\em extended} conditional gradient noise covariance matrix
\begin{align}
\mathcal{R}_{s}(\bm{w}_i) &\triangleq \E\left[\bm{s}_i\bm{s}_i^{\top}| \bm{q}_{0:i-1}, \bm{\psi}_{0:i-1} \right]\nonumber\\
&=\textnormal{blkdiag}\{R_{s,1}(\bm{w}_{1,i-1}), \ldots, R_{s,N}(\bm{w}_{N,i-1}) \}
\label{eq:gradNoiseCovDefLarge},
\end{align}
where the last equality follows from \eqref{eq:gradNoiseUncorrelated}. 

Likewise, we introduce the conditional compression-error covariance matrix for each agent, namely,
\beq 
R_{q,k}(\bm{\delta}_{k,i}) = \E\left[ \bm{e}_{k,i}\bm{e}_{k,i}^{\top}|\bm{\delta}_i\right],
\eeq 
and its extended counterpart 
\begin{align}
& \mathcal{R}_{q}(\bm{\delta}_i) \triangleq \E\left[\bm{e}_i\bm{e}_i^{\top}|\bm{\delta}_i \right] \nonumber \\
& =\textnormal{blkdiag}\{R_{q,1}(\bm{\delta}_{1,i}), \ldots, R_{q,N}(\bm{\delta}_{N,i}) \},
\label{eq:quantNoiseCovDef}
\end{align}
where the last equality holds in view of \eqref{eq:equantzerocorr}.

\section{Proof of Theorem \ref{th:ACTCMSD}}
\label{app:MSDthProof}
\begin{IEEEproof}
The proof consists of three main steps:
\\
\noindent
$(i)$ In Sec.~\ref{app:quantStateApprox} we show that the evolution of the compressed states $\widetilde{\bm{q}}_i$ can be more conveniently replaced by an {\em approximate model} $\bm{q}_{a,i}$.
\\
\noindent
$(ii)$ In Sec.~\ref{sec:agPerfInter} we relate the interactions of the agents' iterates to the correlation matrix of the approximate model $\bm{q}_{a,i}$ to prove \eqref{eq:smallCorrRes}. 
\\
\noindent
$(iii)$ In Sec.~\ref{sec:finalProofStep} we examine the gradient noise and the compression error to obtain an explicit characterization of the trace of the correlation matrix of $\bm{q}_{a,i}$ to prove \eqref{eq:covRecursTraceSubsTh}.


\subsection{Approximate Model}
\label{app:quantStateApprox}
By appealing to a {\em network coordinate transformation}, which is explained in Appendix~\ref{sec:netTrans}, we can represent the evolution of the compressed iterates $\widetilde{\bm{q}}_i$ in terms of a {\em coordinated component} $\bar{\bm{q}}_i$ and an {\em error component} $\widecheck{\bm{q}}_i$.
The stability analysis of the ACTC strategy in~\cite{CarpentieroMattaSayed2022} establishes that the coordinated component $\bar{\bm{q}}_i$ has a dominant role, while the error component $\widecheck{\bm{q}}_i$ acts as a perturbation term --- see Appendix~\ref{app:asymBounds}. 
For these reasons, it is meaningful, in the evolution of $\bar{\bm{q}}_i$ in  \eqref{qTildeSystem}, to focus on the recursion for the coordinated component:
\begin{align}
\bar{\bm{q}}_{i} &= \left( I_M - \mu\,\zeta\, G_{11} \right) \bar{\bm{q}}_{i-1}  - \mu\,\zeta\, G_{12}\widecheck{\bm{q}}_{i-1}\nonumber\\
&+\zeta(\pi^{\top} \otimes I_M)\bm{e}_i - \mu\,\zeta(\pi^{\top} \otimes I_M)\bm{s}_{i},
\label{qaSystempreliminary}
\end{align}
where the matrices $G_{11}$ and $G_{12}$ are defined in \eqref{eq:gDef}.
By neglecting the error component $\widecheck{\bm{q}}_{i-1}$ in \eqref{qaSystempreliminary}, we arrive at the approximate recursion:
\begin{align}
\bm{q}_{a,i} &= \left( I_M - \mu\,\zeta\, G_{11} \right) \bm{q}_{a,i-1} \nonumber\\
&+ \zeta(\pi^{\top} \otimes I_M)\bm{e}_i - \mu\,\zeta (\pi^{\top} \otimes I_M)\bm{s}_{i}.
\label{qaSystem}
\end{align}
We next show that the performance of each individual vector $\bm{q}_{k,i}$ can be approximated through $\bm{q}_{a,i}$. Since we are interested in the mean-square-error performance, one fundamental descriptor will be the correlation matrix of $\bm{q}_{a,i}$:
\beq 
\Pi_{a,i} \triangleq \E\left[ \bm{q}_{a,i} \bm{q}_{a,i}^{\top} \right].
\label{eq:qCovSimp}
\eeq 
Expanding the definition in \eqref{eq:qCovSimp} using \eqref{qaSystem}, exploiting \eqref{eq:equantzeromean} and \eqref{eq:gradNoiseUncorrelated}, and using \eqref{eq:gradNoiseCovDefLarge} and \eqref{eq:quantNoiseCovDef}, we obtain
\begin{align} 
\Pi_{a,i} & = D\,\Pi_{a,i-1} D 
\nonumber\\&+ \mu^2\zeta^2(\pi^{\top} \otimes I_M)\E\left[\mathcal{R}_s(\bm{w}_{i-1})\right](\pi \otimes I_M) \nonumber \\
& + \zeta^2(\pi^{\top} \otimes I_M)\E\left[\mathcal{R}_{q}(\bm{\delta}_i)\right](\pi \otimes I_M),
\label{eq:covComp}
\end{align}
where $D = (I_M - \mu\,\zeta\, G_{11})$ is a real symmetric $M \times M$ matrix.
Note that under mean-square stability of the ACTC strategy, $D$ is Schur-stable~\cite{CarpentieroMattaSayed2022}.
By adding and subtracting the gradient noise covariance matrix \eqref{eq:gradNoiseCovDefLarge} evaluated at the global minimizer $w^o$, we can write \eqref{eq:covComp} as
\begin{align}
&\Pi_{a,i}  = D\,\Pi_{a,i-1} D 
\nonumber\\
&+ \mu^2\zeta^2(\pi^{\top} \otimes I_M)\mathcal{R}_{s}(\mathds{1}_N\otimes w^o)(\pi \otimes I_M) \nonumber \\
& + \zeta^2(\pi^{\top} \otimes I_M)\E\left[\mathcal{R}_{q}(\bm{\delta}_i)\right](\pi \otimes I_M) + \mu^2\zeta^2 \nonumber \\
& \times (\pi^{\top} \otimes I_M)\E\left[\mathcal{R}_{s}(\bm{w}_{i-1}) - \mathcal{R}_{s}(\mathds{1}_N\otimes w^o)\right](\pi \otimes I_M).
\label{eq:covComp2}
\end{align}
The next lemma shows that the last term on the RHS of \eqref{eq:covComp2} can be neglected in the small-$\mu$ regime, such that we can approximate the true correlation matrix $\Pi_i$ through the matrix $\Pi_{a,i}^{o}$ that characterizes the {\em unperturbed} model:
\begin{align} 
\Pi_{a,i}^{o} & = D\,\Pi_{a,i-1}^o D \nonumber\\
&+ \mu^2\zeta^2(\pi^{\top} \otimes I_M)\mathcal{R}_{s}(\mathds{1}_N\otimes w^o)(\pi \otimes I_M) \nonumber \\
& + \zeta^2(\pi^{\top} \otimes I_M)\E\left[\mathcal{R}_{q}(\bm{\delta}_i)\right](\pi \otimes I_M),
\label{eq:covComp3}
\end{align}
in place of the matrix $\Pi_{a,i}$ appearing in \eqref{eq:covComp2}. 
More specifically, the next lemma establishes the equivalence between the original and the unperturbed model by showing that the correlation matrix $\E[\widetilde{\bm{q}}_{\ell,i} \widetilde{\bm{q}}_{k,i}^{\top}]$ between the compressed iterates for any two agents $\ell$ and $k$ (i.e., every $(\ell,k)$ block of the extended-state correlation matrix $\Pi_i$) is equivalent, up to $O(\mu^{3/2})$ perturbations, to $\Pi^{o}_{a,i}$.

\begin{lemma}[{\bf Approximate compressed state model}]
\label{lem:simpModelApprox}
For sufficiently small values of the step-size $\mu$ and of the stability parameter $\zeta$ such that the ACTC strategy is mean-square stable, 
it holds that
\beq
\limsup_{i \rightarrow \infty} \left\| \Pi_i - \mathds{1}_N\mathds{1}_N^{\top} \otimes \Pi_{a,i}^o \right\| = O(\mu^{3/2}).
\label{eq:simpModelApprox}
\eeq
Relation \eqref{eq:simpModelApprox} implies in particular
\beq
\limsup_{i\rightarrow\infty}
\left|
\E[\widetilde{\bm{q}}_{\ell,i}^{\top} \widetilde{\bm{q}}_{k,i}] - \textnormal{Tr}[\Pi_{a,i}^o] 
\right| = O(\mu^{3/2}).
\label{eq:scalarprodapproxappendix}
\eeq
\end{lemma}
\begin{IEEEproof}
See Appendix~\ref{app:simpRecursmModelApprox}.
\end{IEEEproof}

\subsection{Agents' Interaction --- Eq. \eqref{eq:smallCorrRes}}
\label{sec:agPerfInter}
Lemma~\ref{lem:simpModelApprox} reveals that the mutual agents' is captured by $\Pi_{a,i}^o$. 
In particular, we can approximate the scalar product between the compressed iterates of any agents $\ell$ and $k$ (including the case $\ell=k$) through the trace of $\Pi_{a,i}^o$. 
We can readily restate \eqref{eq:scalarprodapproxappendix} in terms of the iterates $\widetilde{\bm{w}}_{k,i}$ in \eqref{eq:centeredwmaintext}. 
In fact, from the combination step \eqref{eq:combStep} we have 
$\widetilde{\bm{w}}_{k,i}=\sum_{\ell=1}^N a_{\ell k} \widetilde{\bm{q}}_{\ell,i}$, yielding
\beq
\widetilde{\bm{w}}_{\ell,i}^{\top} \widetilde{\bm{w}}_{k,i}=
\sum_{j=1}^N \sum_{h=1}^N a_{j \ell} a_{h k} \widetilde{\bm{q}}_{j,i}^{\top} \widetilde{\bm{q}}_{h,i}.
\eeq
Since $A$ is left-stochastic, this implies
\beq
\widetilde{\bm{w}}_{\ell,i}^{\top} \widetilde{\bm{w}}_{k,i} - \textnormal{Tr}[\Pi_{a,i}^o]\!=\!\!
\sum_{j=1}^N \! \sum_{h=1}^N a_{j \ell} a_{h k} \!\left(\widetilde{\bm{q}}_{j,i}^{\top} \widetilde{\bm{q}}_{h,i}
- \textnormal{Tr}[\Pi_{a,i}^o]
\right),
\eeq
and \eqref{eq:scalarprodapproxappendix} implies
\beq
\limsup_{i\rightarrow\infty}
\left|
\E[\widetilde{\bm{w}}_{\ell,i}^{\top} \widetilde{\bm{w}}_{k,i}] - \textnormal{Tr}[\Pi_{a,i}^o] 
\right| = O(\mu^{3/2}).
\label{eq:scalarprodapproxappendixintermsofw}
\eeq
Let now $\rho_{\ell k,i}\triangleq \E \widetilde{\bm{w}}_{\ell,i}^{\top} \widetilde{\bm{w}}_{k,i}$. Since we can write
\beq
\left|
\rho_{\ell k,i} - \rho_{k k,i}
\right|\leq
\left|
\rho_{\ell k,i} - \textnormal{Tr}[\Pi_{a,i}^o]
\right|
+
\left|
\rho_{\ell k,i} - \textnormal{Tr}[\Pi_{a,i}^o]
\right|,
\eeq
from \eqref{eq:scalarprodapproxappendixintermsofw} we obtain
\beq
\limsup_{i\rightarrow\infty}
\left|
\rho_{\ell k,i} - \rho_{k k,i}
\right|=O(\mu^{3/2}).
\label{eq:diffrhoestimate}
\eeq
On the other hand, we have
\begin{align}
&\frac{\rho_{\ell k,i}}{\sqrt{\rho_{\ell \ell,i} \rho_{kk,i}}} - 1
=
\frac{\rho_{\ell k,i} - \rho_{kk,i}}{\sqrt{\rho_{\ell \ell,i} \rho_{kk,i}}} +\frac{\sqrt{\rho_{kk,i}} - \sqrt{\rho_{\ell \ell,i}}}{\sqrt{\rho_{\ell\ell,i}}}
\nonumber\\
&=\frac{\rho_{\ell k,i} - \rho_{kk,i}}{\sqrt{\rho_{\ell \ell,i} \rho_{kk,i}}} + \frac{\rho_{kk,i} - \rho_{\ell\ell,i}}{\sqrt{\rho_{\ell\ell,i}}\left(\sqrt{\rho_{kk,i}} + \sqrt{\rho_{\ell \ell,i}}\right)}.
\label{eq:rhodevelop}
\end{align}
Applying the properties of the limit superior and inferior, from \eqref{eq:rhodevelop} we obtain the following chain of relations:
\begin{align}
&\limsup_{i\rightarrow\infty}\left|\frac{\rho_{\ell k,i}}{\sqrt{\rho_{\ell \ell,i} \rho_{kk,i}}} - 1\right|
\leq\frac{\limsup\limits_{i\rightarrow\infty}\left|\rho_{\ell k,i} - \rho_{kk,i}\right|}{\sqrt{\liminf\limits_{i\rightarrow\infty}\rho_{\ell \ell,i} \,\liminf\limits_{i\rightarrow\infty}\rho_{kk,i}}} \nonumber\\
\nonumber\\ 
&+\frac{\limsup\limits_{i\rightarrow\infty} \left|\rho_{kk,i} - \rho_{\ell \ell,i}\right|}
{\sqrt{\liminf\limits_{i\rightarrow\infty} \rho_{\ell\ell,i}}
\left(\sqrt{\liminf\limits_{i\rightarrow\infty}\rho_{kk,i}} + \sqrt{\liminf\limits_{i\rightarrow\infty}\rho_{\ell \ell,i}}\right)}\nonumber\\
\nonumber \\
&\leq
\frac{O(\mu^{3/2})}{\sqrt{\left(\kappa\mu - O(\mu^{3/2})\right)\left(\kappa\mu - O(\mu^{3/2})\right)}}\nonumber\\
&\!+\!
\frac{O(\mu^{3/2})}{\sqrt{\left(\kappa\mu \!-\! O(\mu^{3/2})\right)}
\!\left(\!\sqrt{\kappa\mu \!-\! O(\mu^{3/2})} \!+\! \sqrt{\kappa\mu \!-\! O(\mu^{3/2})}\right)
}
\nonumber\\
&=O(\mu^{1/2}).
\end{align}
which gives \eqref{eq:smallCorrRes}.

\subsection{MSE bounds --- Eq. \eqref{eq:covRecursTraceSubsTh}}
\label{sec:finalProofStep}

From \eqref{eq:scalarprodapproxappendixintermsofw} we see that to characterize the MSE we need to evaluate the trace of $\Pi_{a,i}^o$ from \eqref{eq:covComp3}. 
For convenience of notation, we introduce the matrices:
\begin{align}
\bar{R}_{s} &= (\pi^{\top} \otimes I_M)\mathcal{R}_{s}(\mathds{1}_N\otimes w^o)(\pi \otimes I_M)\nonumber\\
&= \sum_{k=1}^N \pi^2_k R_{s,k}(w^o), \label{eq:ezNotationGradCov} \\
\bar{R}_{q,i} &= (\pi^{\top} \otimes I_M)\E\left[\mathcal{R}_{q}(\bm{\delta}_i)\right](\pi \otimes I_M) \nonumber\\
&= \sum_{k=1}^N \pi^2_k \E\left[ R_{q,k}(\bm{\delta}_{k,i})\right].
\label{eq:ezNotationQuantCov}
\end{align}
Developing the recursion in \eqref{eq:covComp3} leads to 
\begin{align}
\Pi_{a,i}^{o} &=  D^i \Pi_{a,0}^o D^i + \mu^2\zeta^2 \sum_{j=0}^{i} D^j \bar{R}_{s} D^j
\nonumber\\
&+ \zeta^2 \sum_{j=0}^{i} D^j \bar{R}_{q,i-j} D^j.
\label{eq:devCovRecurs}
\end{align}
The first term on the RHS vanishes as $i\rightarrow\infty$ since $D$ is Schur-stable. We now characterize the other two terms.

\subsubsection{Gradient Noise Steady-State}
For the second term on the RHS of  \eqref{eq:devCovRecurs} we can write
\begin{align}
\textnormal{Tr}\left[ \sum_{j=0}^{\infty} D^j \bar{R}_{s} D^j \right] & =  \textnormal{Tr}\left[ \sum_{j=0}^{\infty} D^{2j} \bar{R}_{s} \right] \nonumber \\
& = \textnormal{Tr}\left[ (I_M - D^2)^{-1}\bar{R}_{s}\right]
\label{eq:gradNoiseTrace}
\end{align}
where the first equality follows from the cyclic property of the trace operator, while the second equality follows by computing the geometric matrix series of the Schur-stable matrix $D$. Recalling the definition of $D$ in \eqref{eq:dDef}:
\begin{align} 
I_M-D^2 & =(I_M+D)(I_M-D) \nonumber \\
& = (2 I_M - \mu\,\zeta\, G_{11})\, \mu\,\zeta\, G_{11}, 
\end{align}
and then 
\begin{align}
&(I_M - D^2)^{-1}  = \left(I_M - \mu\,\zeta\frac{G_{11}}{2} \right)^{-1}\frac{G_{11}^{-1}}{2\,\mu\,\zeta} \nonumber \\
& = \frac{G_{11}^{-1}}{2\,\mu\,\zeta} \!+\! \left[  \left(I_M - \mu\,\zeta\frac{G_{11}}{2} \right)^{-1} - I_M\right]\frac{G_{11}^{-1}}{2\,\mu\,\zeta},
\label{eq:DInvNice}
\end{align}
where in the second equality we added and subtracted the identity matrix $I_M$. Regarding the second term on the RHS of \eqref{eq:DInvNice}, we have that:
\begin{align}
& \left[  \left(I_M - \mu\,\zeta\frac{G_{11}}{2} \right)^{-1} - I_M\right]\frac{G_{11}^{-1}}{2\,\mu\,\zeta}
\nonumber \\
& = \left[\sum_{j=1}^{\infty} \left(\mu\,\zeta\frac{G_{11}}{2} \right)^j \right] \frac{G_{11}^{-1}}{2\,\mu\,\zeta}  = \frac{1}{4}\left[\sum_{j'=0}^{\infty} \left(\mu\,\zeta\frac{G_{11}}{2} \right)^{j'} \right]  \nonumber \\
& = \frac{1}{4}\left(I_M - \mu\,\zeta\frac{G_{11}}{2}\right)^{-1}. 
\label{eq:pertMatGradNoise}
\end{align}
Since the trace and the inverse are continuous operators, we know that as $\mu\rightarrow 0$ the trace of  $(I_M - \mu\,\zeta G_{11}/2)^{-1} \bar{R}_{s}$ converges to $\textnormal{Tr}[\bar{R}_{s}]$, which in particular implies that:
\beq
\textnormal{Tr}\left[\left(I_M - \mu\,\zeta\frac{G_{11}}{2}\right)^{-1} \bar{R}_{s}\right]=O(1).
\eeq
We can therefore write \eqref{eq:gradNoiseTrace} as
\begin{align}
& \textnormal{Tr}\left[ \sum_{j=0}^{\infty} D^j \bar{R}_{s} D^j \right] =
\frac{1}{2\mu\zeta}\textnormal{Tr}\left[ G_{11}^{-1}\,\bar{R}_{s}\right]
+ O(1)\nonumber\\
& =\frac{1}{4\,\mu\,\zeta} \textnormal{Tr}\left[ \left( \sum_{k=1}^N \alpha_k \pi_k R_{u,k}\right)^{-1}\left( \sum_{k=1}^N \pi_k^2R_{s,k}(w^o)\right) \right]\nonumber\\
&+ O(1),
\label{eq:finalGradNoiseCovTrace}
\end{align}
where the first equality follows by \eqref{eq:gradNoiseTrace} after substituting \eqref{eq:pertMatGradNoise} into \eqref{eq:DInvNice}, while the second equality follows from \eqref{eq:ezNotationGradCov} and \eqref{eq:g11DefEz}.

\subsubsection{Compression Error Steady-State}
\label{sec:quantnoiseanalysislemma2main}
Substituting \eqref{eq:ezNotationQuantCov} into the third term on the RHS of \eqref{eq:devCovRecurs} we have
\begin{align}
& \textnormal{Tr}\left[ \sum_{j=0}^{i} D^j \bar{R}_{q,i-j} D^j \right] \nonumber \\
& = \textnormal{Tr}\left[ \sum_{j=0}^i D^j\left( \sum_{k=1}^N \pi^2_k \E\left[ R_{q,k}(\bm{\delta}_{k,i-j})\right] \right)D^j \right] \nonumber \\
& \overset{(a)}{=} \sum_{j=0}^i \sum_{k=1}^N \pi^2_k \, \textnormal{Tr}\left[ D^{2j} \E\left[ R_{q,k}(\bm{\delta}_{k,i-j}) \right] \right]
\nonumber \\
& \overset{(b)}{\leq} \sum_{j=0}^i \sum_{k=1}^N \pi^2_k 
\sum_{m=1}^M \lambda_m\big(D^{2j}\big)  \lambda_m\big(R_{q,k}(\bm{\delta}_{k,i-j})\big) 
\nonumber \\
& \overset{(c)}{\leq} \sum_{j=0}^i (1 - \mu\,\zeta\,\nu)^{2j}
\sum_{k=1}^N \pi^2_k 
\underbrace{\sum_{m=1}^M \lambda_m\big(R_{q,k}(\bm{\delta}_{k,i-j})\big)}_{=\textnormal{Tr}[R_{q,k}(\bm{\delta}_{k,i-j})]} 
\nonumber \\
& \overset{(d)}{\leq} \sum_{j=0}^i (1 - \mu\,\zeta\,\nu)^{2j} \sum_{k=1}^N \pi_k^2 \omega_k \E\| \bm{\delta}_{k,i-j}\|^2 \nonumber \\
& \overset{(e)}{=} \sum_{j=0}^{i}(1-\mu\,\zeta\,\nu)^{2j}\E\|\bm{\delta}_{i-j}\|^2_{\Omega_{\pi}},
\label{eq:quantNoiseCovTrace}
\end{align}
where step $(a)$ follows from linearity and the cyclic property of the trace operator and step $(b)$ follows from Von Neumann's trace inequality (with singular values replaced by eigenvalues, since the involved matrices are symmetric), where we denoted by $\lambda_m(\cdot)$ the $m$-th eigenvalue of its matrix argument~\cite{Bhatia}.
Step $(c)$ follows by observing that the diagonal entries of $D^j$ are upper bounded by the maximum eigenvalue of $D^j$ (recall that $D$ is positive definite), and that the maximum eigenvalue is in turn upper bounded by $(1 - \mu\,\zeta\,\nu)^j$ in view of \eqref{eq:IminusG11TimeBound}. 
Step $(d)$ follows from the variance bound in \eqref{eq:equantvariancebound}.
In order to obtain step $(e)$, we recall that given an $MN\times 1$ block vector $x = \textnormal{col}\{x_1, x_2, \ldots, x_N\}$, with $M \times 1$ entries $x_k$, and a set of positive scalars $\{y_k\}_{k=1}^N$, it holds that 
\beq  
\sum_{k=1}^N y_k \|x_k\|^2 = x^{\top} Y x = \|x\|^2_Y,
\label{eq:auxiliaryResNorm}
\eeq
where $Y = \textnormal{blkdiag}\left\{ y_1 I_M, y_2 I_M, \ldots, y_N I_M\right\}$. Computing the inner summation over $k$ in \eqref{eq:quantNoiseCovTrace}, and since the $MN\times 1$ extended vector $\bm{\delta}_i$ stacks the $M\times 1$ vectors $\bm{\delta}_{k,i}$, we can introduce the diagonal weighting matrix 
\beq 
\Omega_{\pi} = \textnormal{blkdiag}\left\{ \pi_1^2\omega_1 I_M, \ldots, \pi_N^2\omega_N I_M\right\},
\label{eq:omegaPiDef}
\eeq 
and get from \eqref{eq:auxiliaryResNorm}:
\beq
\E\|\bm{\delta}_{i-j}\|^2_{\Omega_{\pi}}=\sum_{k=1}^N \pi_k^2 \omega_k \E\| \bm{\delta}_{k,i-j}\|^2.
\eeq
In order to characterize the last summation in \eqref{eq:quantNoiseCovTrace}, we take advantage of the asymptotic behavior of the weighted norm of the differential iterates.

\begin{lemma}[{\bf Weighted energy of differential iterates}]\label{lem:deltaApprox}
For sufficiently small values of the step-size $\mu$ and of the stability parameter $\zeta$ such that the ACTC strategy is mean-square stable, the $\Omega_{\pi}$-weighted norm of the differential iterates $\bm{\delta}_i$
 satisfies the bound
\begin{align}
\limsup_{i \rightarrow \infty} \E\|\bm{\delta}_i\|^2_{\Omega_{\pi}} & \leq \mu^2\sum_{k=1}^N \pi_k^2\omega_k\textnormal{Tr}[R_{s,k}(w^o)] \nonumber \\
& + \mu^2\zeta\,c + O(\mu^{3}),
\label{eq:finalDeltaExpressionLemma}
\end{align}
where $c$ is a constant independent of $\mu$ and depending on the compression parameters $\{ \omega_k\}_{k=1}^N$.
\begin{IEEEproof}
See appendix~\ref{app:weightedDiff}.
\end{IEEEproof}
\end{lemma}

Next we observe that if a sequence $\{z_i\}$ has finite limit superior equal to $z$, it holds that (see Lemma~\ref{lem:geomSum})
\beq 
\limsup_{i\rightarrow \infty} \sum_{j=0}^i \beta^j z_{i-j} \leq \frac{z}{1 - \beta},\qquad (0<\beta<1).
\label{eq:geomSumPiezzomaintext}
\eeq 
We now apply \eqref{eq:geomSumPiezzomaintext} to the last summation in \eqref{eq:quantNoiseCovTrace} with the choices:
\beq
z_i=\E\|\bm{\delta}_{i}\|^2_{\Omega_{\pi}},\qquad \beta = (1 - \mu\,\zeta\,\nu)^2.
\eeq
By further using \eqref{eq:finalDeltaExpressionLemma} to upper bound the limit superior of $z_i$, we finally obtain: 
\begin{align}
& \limsup_{i \rightarrow \infty} \textnormal{Tr}\left[ \sum_{j=0}^{i} D^j \bar{R}_{q,i-j} D^j \right] \nonumber \\ 
& \leq \frac{\mu}{\zeta\,\nu\,(2-\mu\,\zeta\,\nu)}
\sum_{k=1}^N \pi_k^2\omega_k \textnormal{Tr}[R_{s,k}(w^o)]
\nonumber \\
& + \frac{\mu\,c}{\nu\,(2 - \mu\,\zeta\,\nu)}  + O(\mu^{2}) \nonumber \\
& \leq \frac{\mu}{2\,\zeta\,\nu}
\sum_{k=1}^N \pi_k^2\omega_k \textnormal{Tr}[R_{s,k}(w^o)] 
+ \frac{\mu\,c}{2\,\nu} + O(\mu^{2}),
\label{eq:finalQuantNoiseCovTrace}
\end{align}
where in the second inequality we used the asymptotic expansion $1/(2 - \mu\,\zeta\,\nu) = 1/2 + O(\mu)$.

Applying \eqref{eq:finalGradNoiseCovTrace} and \eqref{eq:finalQuantNoiseCovTrace} in \eqref{eq:devCovRecurs}, and further exploiting definitions \eqref{eq:gradnoisetermnewdef} and \eqref{eq:compnoisetermnewdef}, we can write:
\begin{align}
\mu\,\Delta_s - O(\mu^{2})
&\leq
\liminf_{i\rightarrow\infty}\textnormal{Tr}[\Pi_{a,i}^o]\nonumber\\
&\leq\limsup_{i\rightarrow\infty}\textnormal{Tr}[\Pi_{a,i}^o]\nonumber\\
&\leq
\mu\left(\Delta_s + \Delta_{\omega}\right) + O(\mu^{2}),
\end{align}
and the proof is complete due to \eqref{eq:scalarprodapproxappendixintermsofw}.
\end{IEEEproof}

\section{Proof of Property \ref{prop:gradNoise}}
\label{app:gradNoiseProp}
\begin{IEEEproof}
The exact expression of the gradient noise can be obtained by substituting \eqref{eq:realGrad} and \eqref{eq:instGradApprox} into \eqref{eq:scaledGradNoiseDef}, with the following algebraic manipulations:
\begin{align}
&\bm{s}_{k,i} 
= 2\alpha_k\Big\{
\bm{u}_{k,i}\left(\bm{u}_{k,i}^{\top}\bm{w}_{k,i-1} - \bm{d}_{k,i} \right) 
\nonumber\\
&- 
\left( R_{u,k}\bm{w}_{k,i-1} - r_{du,k} \right) 
 \Big\}
 \nonumber \\
& \overset{(a)}{=} 2\alpha_k
\Big\{
(\bm{u}_{k,i}\bm{u}_{k,i}^{\top} \!-\!  R_{u,k})\bm{w}_{k,i-1}  
- \bm{u}_{k,i}\bm{d}_{k,i} \!+\! R_{u,k} w^o 
\Big\}
\nonumber \\
& \overset{(b)}{=} 2\alpha_k
\Big\{
\bm{U}_{k,i}\bm{w}_{k,i-1}  
 - \bm{u}_{k,i}(\bm{u}_{k,i}^{\top}w^o + \bm{v}_{k,i}) + R_{u,k} w^o \Big\}
 \nonumber \\
& = 2\alpha_k\Big\{\bm{U}_{k,i}(\bm{w}_{k,i-1} - w^o) - \bm{u}_{k,i}\bm{v}_{k,i}\Big\},
\label{eq:gradNoiseExprDer}
\end{align} 
where in step $(a)$ we used the fact that setting \eqref{eq:realGrad} equal to zero we end up with the relation 
\beq
R_{u,k}w^o = r_{du,k},
\label{eq:w0asafunofcov}
\eeq
and in step $(b)$ we invoked \eqref{eq:linRegModel} and used the definition $\bm{U}_{k,i} = \bm{u}_{k,i}\bm{u}_{k,i}^{\top} -  R_{u,k}$. It follows that
\begin{align}
&\E[\bm{s}_{k,i}|\bm{q}_{0:i-1} , \bm{\psi}_{0:i-1}]=2\,\alpha_k
\nonumber\\
&\times\Big\{
\E[\bm{U}_{k,i}(\bm{w}_{k,i-1} - w^o)|\bm{q}_{0:i-1} , \bm{\psi}_{0:i-1}] \nonumber \\
& \quad - \E[\bm{u}_{k,i}\bm{v}_{k,i}|\bm{q}_{0:i-1} , \bm{\psi}_{0:i-1}]
\Big\}\nonumber\\
&=\E[\bm{U}_{k,i}](\bm{w}_{k,i-1} - w^o) - \E[\bm{u}_{k,i} \bm{v}_{k,i}]=0,
\end{align}
where in the last equality conditioning is removed because the pair $(\bm{u}_{k,i},\bm{v}_{k,i})$ is statistically independent from all the past variables up to instant $i-1$ (since the past variables depend on the past noises and regressors and by assumption the sequences $\bm{v}_{k,i}$ and $\bm{u}_{k,i}$ are independent over time). 
The last equality follows by observing that the regressors $\bm{u}_{k,i}$ and the zero-mean noise $\bm{v}_{k,i}$ are statistically independent, so that $\E[\bm{U}_{k,i}] = \E[\bm{u}_{k,i}\bm{u}_{k,i}^{\top}] -  R_{u,k}=0$.
Using similar arguments, it is straightforward to prove \eqref{eq:gradNoiseUncorrelated} (by further exploiting independence across agents) and \eqref{eq:gradNoiseCovIntro}. 

To complete the proof we need to establish \eqref{eq:covdifflimsupmain}.
To this end, we use \eqref{eq:gradNoiseSingleAgentCovLimit} to obtain: 
\begin{align}
& \|R_{s,k}(w) - R_{s,k}(w^o)\| \nonumber\\
&= \|\E\left[\bm{U}_{k,i}(w-w^o)(w -w^o)^{\top}\bm{U}_{k,i}^{\top} \right]\| \nonumber \\
& \overset{(a)}{\leq} \E\| \bm{U}_{k,i}\|^2\,\|(w - w^o) (w - w^o)^{\top}\|\nonumber\\
& \overset{(b)}{\leq} \E\| \bm{U}_{k,i}\|^2\,\textnormal{Tr}[(w - w^o) (w - w^o)^{\top}]\nonumber\\
& = \E\| \bm{U}_{k,i}\|^2\, \|w - w^o\|^2,
\label{eq:niceResultOnGradNoise}
\end{align}
where in step $(a)$ we apply Jensen's inequality to the convex function $\|\cdot\|$ and use submultiplicativity of the operator norm, whereas in step $(b)$ we apply the property $\|X\| \leq \textnormal{Tr}[X]$ holding for any (symmetric) positive semidefinite matrix $X$.
Setting $w = \bm{w}_{k,i-1}$ in \eqref{eq:niceResultOnGradNoise} and taking expectation, we get
\begin{align}
\E\|R_{s,k}(\bm{w}_{k,i-1}) &- R_{s,k}(w^o)\|\nonumber\\ 
&\leq \E\| \bm{U}_{k,i}\|^2\,\E\|\bm{w}_{k,i-1} - w^o \|^2,  
\label{eq:higherTermBoring}
\end{align}
Since from~\cite{CarpentieroMattaSayed2022} we know that $\limsup_{i\rightarrow\infty}\E\|\widetilde{\bm{w}}_{k,i-1}\|^2 = O(\mu)$, Eq. \eqref{eq:higherTermBoring} implies the claim in \eqref{eq:covdifflimsupmain}.

Finally, we show that \eqref{eq:covdifflimsupmain} implies \eqref{eq:meansquaregradmaintheorem}. 
Exploiting the inequality $|\textnormal{Tr}[X]|\leq M \|X\|$ (which can be seen as a corollary of Von Neumann's trace inequality~\cite{Bhatia}), we have:
\begin{align}
&-M \|R_{s,k}(\bm{w}_{k,i-1})]  - R_{s,k}(w^o)\|\nonumber\\
&\leq\textnormal{Tr}[R_{s,k}(\bm{w}_{k,i-1}) - R_{s,k}(w^o)]\nonumber\\
&\leq
M \|R_{s,k}(\bm{w}_{k,i-1})]  - R_{s,k}(w^o)\|.
\label{eq:tracenormRsbound}
\end{align}
Observe now that, by taking expectation in \eqref{eq:gradNoiseCovIntro}, we have: 
\beq
\E\|\bm{s}_{k,i}\|^2 = \E\,\textnormal{Tr}[R_{s,k}(\bm{w}_{k,i-1})],
\eeq
which, when used in \eqref{eq:tracenormRsbound}, yields \eqref{eq:meansquaregradmaintheorem}.
\end{IEEEproof}

\section{Collection of Useful Background Results}
\label{app:collection}

\subsection{Network Error Dynamics} 
\label{app:collectionA}
The mean-square-performance of the ACTC diffusion strategy can be more conveniently studied in terms of the centered variables
\beq 
\widetilde{\bm{w}}_{k,i} \triangleq \bm{w}_{k,i} - w^o, \quad \widetilde{\bm{\psi}}_{k,i} \triangleq \bm{\psi}_{k,i} - w^o, \quad
\widetilde{\bm{q}}_{k,i}  \triangleq \bm{q}_{k,i} - w^o,
\eeq 
which, by substituting into \eqref{ACTC}, yield
\begin{subequations}\label{centeredACTC}
    \begin{align}
      & \widetilde{\bm{\psi}}_{k,i} = (I_M - \mu\, H_{k})\widetilde{\bm{w}}_{k,i-1} - \mu \,\bm{s}_{k,i}\\
      & \widetilde{\bm{q}}_{k,i} = \widetilde{\bm{q}}_{k,i-1} + \zeta \,\bm{Q}_k(\widetilde{\bm{\psi}}_{k,i} - \widetilde{\bm{q}}_{k,i-1}) \\
      & \widetilde{\bm{w}}_{k,i} = \sum_{\ell \in \mathcal{N}_k} a_{\ell k}\widetilde{\bm{q}}_{\ell,i}
      \label{eq:centeredACTCcomb}
    \end{align}
\end{subequations} 
where we introduced the scaled-Hessian matrix
\beq   
H_k = \alpha_k \nabla^2 J_k(w) =2 \, \alpha_k R_{u,k}, 
\eeq 
and, in the first step, we used \eqref{eq:realGrad}, \eqref{gradNoiseDef}, \eqref{eq:scaledGradNoiseDef}, and \eqref{eq:w0asafunofcov}.
An equivalent formulation of \eqref{centeredACTC} is
\begin{subequations}
\begin{align}
& \bm{\delta}_{i} = [(I_{MN} - \mu\,\mathcal{H})\,\mathcal{A}^{\top}  -  I_{MN}]\widetilde{\bm{q}}_{i-1} 
- \mu\,\bm{s}_{i}
\label{eq:netACTC1}\\
& \widetilde{\bm{q}}_{i} = \widetilde{\bm{q}}_{i-1} + \zeta\,\bm{\mathcal{Q}}(\bm{\delta}_i)
\label{eq:netACTC2}
\end{align}
\end{subequations}
where we used the $MN\times 1$ vectors introduced in Appendix~\ref{app:firstmainapp} in the main manuscript and defined the extended matrices:
\beq
\mathcal{A} \triangleq A \otimes I_M,~~
\mathcal{H} \triangleq \textnormal{blkdiag}\{ H_1, H_2, ..., H_N\}.
\label{hCal}
\eeq
Adding and subtracting $\zeta\,\bm{\delta}_i$ to the RHS of \eqref{eq:netACTC2} gives
\beq
\widetilde{\bm{q}}_i = \widetilde{\bm{q}}_{i-1} + \zeta\,\bm{\delta}_i +  \zeta\,\bm{e}_i,
\label{eq:recallACTCquantState2}
\eeq
where we used the definition of the compression error in \eqref{eq:quantnoisemaintextapp}.
Substituting \eqref{eq:netACTC1}
into \eqref{eq:recallACTCquantState2} leads to
\begin{align}
\widetilde{\bm{q}}_i & = \underbrace{\left\{I_{MN} + \zeta\left[(I_{MN} - \mu\mathcal{H})\mathcal{A}^{\top} - I_{MN}\right]\right\}}_\text{$\mathcal{B}$}\widetilde{\bm{q}}_{i-1} \nonumber \\ 
& + \zeta\,\bm{e}_i - \mu\,\zeta\,\bm{s}_{i}.
\label{eq:qTildeTotal}
\end{align}
Recursion \eqref{eq:qTildeTotal} describes the evolution of the compressed states of the ACTC strategy, from which the iterates $\widetilde{\bm{w}}_{k,i}$ can be obtained from the combination step in \eqref{eq:centeredACTCcomb}. 

\subsection{Correlation Matrix of Compressed Iterates}
The characterization of the mean-square-error performance is based on the analysis of the correlation matrix of the (centered) compressed iterates $\widetilde{\bm{q}}_i$, namely,
\beq 
\Pi_i \triangleq \E\left[ \widetilde{\bm{q}}_i \widetilde{\bm{q}}_i^{\top} \right].
\label{eq:qCov}
\eeq 
This matrix is useful to characterize the mean-square-error performance relative to any pair of agents $\ell$ and $k$, since we can write
\begin{align}
\E\left[ \widetilde{\bm{q}}_{\ell,i}^{\top} \widetilde{\bm{q}}_{k,i} \right] & 
\overset{(a)}{=} \E\left[ \widetilde{\bm{q}}_i^{\top} (E_{\ell k} \otimes I_M) \widetilde{\bm{q}}_i \right] \nonumber \\
& \overset{(b)}{=} \E\left\{\textnormal{Tr}\left[ \widetilde{\bm{q}}_i \widetilde{\bm{q}}_i^{\top}(E_{\ell k} \otimes I_M) \right] \right\} \nonumber \\
& \overset{(c)}{=} \textnormal{Tr}\left\{ \E\left[\widetilde{\bm{q}}_i \widetilde{\bm{q}}_i^{\top} \right](E_{\ell k} \otimes I_M) \right\} \nonumber \\
& \overset{(d)}{=} \textnormal{Tr}\left\{ \Pi_i(E_{\ell k} \otimes I_M) \right\},
\label{eq:selectionMatrix}
\end{align}
where $E_{\ell k}$ is an $N \times N$ selection matrix whose entries are all zeros except for the $(\ell, k)$ element. 
In \eqref{eq:selectionMatrix}, step $(a)$ follows by applying the definition of $E_{\ell k}$; step $(b)$ is a consequence of the equality:
\beq
y^{\top} X y=\textnormal{Tr}(y y^{\top} X)
\eeq
holding for any vector $y$ and matrix $X$ of compatible sizes; step $(c)$ follows from the linearity of the trace and expectation operators; and $(d)$ follows from definition \eqref{eq:qCov}.
Note that for $\ell = k$ we can extract from \eqref{eq:selectionMatrix} the mean-square-error of agent $k$. 

\subsection{Network Coordinate Transformation}
\label{sec:netTrans}
The mean-square analysis of the ACTC diffusion strategy in \cite{CarpentieroMattaSayed2022} was performed by applying a coordinate transformation to the estimated iterates --- see Sec.~V.B therein, based on the Jordan canonical decomposition~\cite{Johnson-Horn} of the transposed combination matrix:
\beq
A^{\top} \triangleq V^{-1} J_{\rm{tot}} V,
\label{aJordan}
\eeq
where
\beq
V =         
\begin{bmatrix}
\pi^{\top}
\\
V_R
\end{bmatrix}
,~~~~~~
V^{-1} = 
\begin{bmatrix}
\mathds{1}_N & V_L 
\end{bmatrix}
,~~~~~~
J_{\rm{tot}} = \begin{bmatrix}
1 & 0
\\
0 & J
\end{bmatrix}.
\label{jordan}
\eeq
The network coordinate transformation is thus
\beq 
\mathcal{V} \triangleq V \otimes I_M = \begin{bmatrix}
\pi^{\top} \otimes I_M \\ V_R \otimes I_M
\end{bmatrix}.
\label{eq:netTransfo}
\eeq 
Recalling Appendix~A.A from \cite{CarpentieroMattaSayed2022}, the matrix $J_{\rm{tot}}$ can be represented as
\beq 
J_{\rm{tot}} = \textnormal{blkdiag}\{J_1,J_2, \ldots, J_{B}\},
\eeq
where each (Jordan shaped) block is 
\beq
J_{n} \triangleq \begin{bmatrix}
\lambda_n & 1 & \\
& \ddots & \ddots & \\
& & \ddots & 1 \\
& & & \lambda_n
\end{bmatrix} =
\lambda_n I_{L_n} + \Theta_{L_n},
\label{eq:JnDef}
\eeq
having denoted by $\lambda_n$ the $n$-th eigenvalue of matrix $A$\footnote{We assume, without loss of generality, that the eigenavalues are sorted in descending order of magnitude.}, by $L_n$ the dimension of the $n$-th block, and by $\Theta_{L_n}$ a square matrix of size $L_n$ that has all zero entries, but for the first diagonal above the main diagonal, which has entries equal to $1$.
Exploiting \eqref{eq:JnDef}, we can write
\beq
J=\Lambda + \Theta.
\label{eq:Jordanrep}
\eeq 
where 
\beq
\Lambda\triangleq \textnormal{blkdiag}\{\lambda_2 I_{L_2},\lambda_3 I_{L_3},\ldots,\lambda_B I_{L_B}\},
\label{eq:Lambdablocks}
\eeq
and
\beq
\Theta\triangleq \textnormal{blkdiag}\{\Theta_{L_2},\Theta_{L_3},\ldots,\Theta_{L_B}\}.
\label{eq:Ublocks}
\eeq
In the following derivations we will also employ the extended matrices:
\beq 
\mathcal{J}_{\rm{tot}} = J_{\rm{tot}} \otimes I_M = \begin{bmatrix}
I_M & 0 \\
0 & \mathcal{J}
\end{bmatrix}, 
\quad \mathcal{J} = J \otimes I_M.
\label{eq:JtotStruct}
\eeq 

\subsection{Representation in Transformed Network Coordinates}
\label{app:usefulrep}
By calling upon the network coordinate transformation in \eqref{eq:netTransfo}, we examine the structure of the matrix $ \mathcal{B}$ in \eqref{eq:qTildeTotal}. Recalling Eq.~(117) from \cite{CarpentieroMattaSayed2022}:
\beq
(I_{MN} - \mu\,\mathcal{H})\mathcal{A}^\top-I_{MN}
= \mathcal{V}^{-1}\,(\mathcal{J}_{\rm{tot}} - \mu\,\mathcal{G} - I_{MN})\,\mathcal{V},
\label{eq:decomposedBigDrivingMatrix}
\eeq
where 
\beq
\mathcal{G} \triangleq \mathcal{V}\,\mathcal{H}\mathcal{A}^\top\,\mathcal{V}^{-1}=
\begin{bmatrix} 
G_{11} & G_{12}\\
G_{21} & G_{22}
\end{bmatrix} 
\label{eq:gDef}, 
\eeq
and the matrices therein were examined in Appendix~A.C in~\cite{CarpentieroMattaSayed2022}. 
In preparation for the forthcoming proofs, we report the explicit form of $G_{11}$, borrowed from~\cite{CarpentieroMattaSayed2022}:
\beq 
G_{11} = \sum_{k=1}^N \pi_k H_k = 2\sum_{k=1}^N \alpha_k\pi_k R_{u,k}.
\label{eq:g11DefEz}
\eeq 
Using \eqref{eq:decomposedBigDrivingMatrix} the matrix $\mathcal{B}$ can be written as
\beq 
\mathcal{B} = \mathcal{V}^{-1}\,\left[I_{MN} + \zeta(\mathcal{J}_{\rm{tot}} - \mu\,\mathcal{G} - I_{MN})\right]\,\mathcal{V}.
\label{eq:bCalDecomp}
\eeq 
Inspecting the structure of the matrices $\mathcal{J}_{\rm{tot}}$ from \eqref{eq:JtotStruct} and $\mathcal{G}$ in \eqref{eq:gDef} we have
\begin{align}
& \mathcal{B} =  \mathcal{V}^{-1}\begin{bmatrix}
I_M - \mu\,\zeta\,G_{11} & -\mu\,\zeta\,G_{12}  \\
-\mu\,\zeta\,G_{21} & \mathcal{J}_{\zeta}-\mu\,\zeta\,G_{22}
\end{bmatrix}\mathcal{V},
\label{eq:bCalDecompBlocks}
\end{align} 
where 
\beq
\mathcal{J}_{\zeta} \triangleq (1-\zeta)I_{M(N-1)} + \zeta\,\mathcal{J},
\eeq 
which by substituting into \eqref{eq:qTildeTotal} after applying the network coordinate transformation in \eqref{eq:netTransfo} leads to the transformed compressed states:
\begin{align}
\widehat{\bm{q}}_i & = 
\begin{bmatrix}
\bar{\bm{q}}_{i}\\
\widecheck{\bm{q}}_{i}
\end{bmatrix}
= \begin{bmatrix}
(\pi^{\top} \otimes I_M)\widetilde{\bm{q}}_{i-1} \\
(V_R \otimes I_M)\widetilde{\bm{q}}_{i-1}
\end{bmatrix} \nonumber \\
&=
\begin{bmatrix}
I_M - \mu\,\zeta\,G_{11} & -\mu\,\zeta\, G_{12}\\
-\mu\,\zeta\,G_{21} & \mathcal{J}_{\zeta}-\mu\,\zeta\,G_{22}
\end{bmatrix} 
\begin{bmatrix}
\bar{\bm{q}}_{i-1} \\
\widecheck{\bm{q}}_{i-1}
\end{bmatrix} \nonumber \\
& + \zeta\begin{bmatrix}
\bar{\bm{e}}_i \\ \widecheck{\bm{e}}_i
\end{bmatrix}
-\mu\,\zeta
\begin{bmatrix}
\bar{\bm{s}}_{i} \\
\widecheck{\bm{s}}_{i}
\end{bmatrix}
.
\label{qTildeSystem}
\end{align}
The $M \times 1$ vector $\bar{\bm{q}}_{i}$ represents the \emph{coordinated component} of the network evolution, and the $M(N-1) \times 1$ vector $\widecheck{\bm{q}}_{i}$ represents the \emph{network error component}. In \eqref{qTildeSystem} we also introduced the transformed versions of the gradient noise $\bm{s}_i$ and compression error $\bm{e}_i$, namely
\begin{align}
\widehat{\bm{s}}_i \!=\! \begin{bmatrix}
\bar{\bm{s}}_i \\ \widecheck{\bm{s}}_i
\end{bmatrix} 
 \!\!=\!\! \begin{bmatrix}
(\pi^{\top} \otimes I_M)\bm{s}_{i} \\
(V_R \otimes I_M)\bm{s}_{i}
\end{bmatrix}, 
\widehat{\bm{e}}_i \!=\! \begin{bmatrix}
\bar{\bm{e}}_i \\ \widecheck{\bm{e}}_i
\end{bmatrix} \!\!=\!\! \begin{bmatrix}
(\pi^{\top} \otimes I_M)\bm{e}_{i} \\
(V_R \otimes I_M)\bm{e}_{i}
\end{bmatrix}.\nonumber\\
\label{eq:transNoises}
\end{align}
Using \eqref{eq:decomposedBigDrivingMatrix}, the evolution of the differential iterates in \eqref{eq:netACTC1} can be written after the application of the network coordinate transformation as 
\beq 
\widehat{\bm{\delta}}_i = (\mathcal{J}_{\rm{tot}} - I_{MN} - \mu\,\mathcal{G})\,\widehat{\bm{q}}_{i-1} - \mu\,\widehat{\bm{s}}_i,
\label{eq:otherDeltaTransRep}
\eeq 
and equivalently, from \eqref{eq:JtotStruct} and \eqref{eq:gDef} (see also Appendix~A.C in \cite{CarpentieroMattaSayed2022}), as
\begin{align}
\widehat{\bm{\delta}}_i & = 
\begin{bmatrix}
\bar{\bm{\delta}}_{i}\\
\widecheck{\bm{\delta}}_{i}
\end{bmatrix}
= \begin{bmatrix}
(\pi^{\top} \otimes I_M)\widetilde{\bm{q}}_{i-1} \\
(V_R \otimes I_M)\widetilde{\bm{q}}_{i-1}
\end{bmatrix} \nonumber \\
&=
\begin{bmatrix}
 - \mu\,G_{11} & -\mu\,G_{12}\\
-\mu\,G_{21} &  \mathcal{J}-I_{M(N-1)}-\mu\,G_{22}
\end{bmatrix} 
\begin{bmatrix}
\bar{\bm{q}}_{i-1} \\
\widecheck{\bm{q}}_{i-1}
\end{bmatrix} \!-\! \mu
\begin{bmatrix}
\bar{\bm{s}}_{i} \\
\widecheck{\bm{s}}_{i}
\end{bmatrix}.
\label{eq:deltaSystem}
\end{align}
Finally, we introduce the following matrix that we use repeatedly in the forthcoming derivations:
\beq 
D \triangleq I_M - \mu\,\zeta\,G_{11},
\label{eq:dDef}
\eeq 
which, in view of \eqref{eq:g11DefEz}, is a real symmetric matrix. For sufficiently small values of the step-size $\mu$ and of the stability parameter $\zeta$, i.e., when the ACTC strategy is mean-square stable, the matrix $D$ is Schur-stable --- see Eq.~(184) in \cite{CarpentieroMattaSayed2022}, and we have that
\beq 
\|I_M - \mu\,\zeta\,G_{11}\| \leq 1 - \mu\,\zeta\,\nu < 1. 
\label{eq:IminusG11TimeBound}
\eeq

\subsection{MSE Bounds for Compressed Iterates}
\label{app:asymBounds}
Theorem 1 in~\cite{CarpentieroMattaSayed2022} shows that for sufficiently small values of $\mu$ and $\zeta$ the ACTC diffusion is mean-square stable. In particular, the two components of the transformed compressed iterates $\widehat{\bm{q}}_i$ in \eqref{qTildeSystem} satisfy the bounds~\cite[Eqs.~(271) and (273)]{CarpentieroMattaSayed2022}:
\beq 
\limsup_{i \rightarrow \infty} \E\|\bar{\bm{q}}_i\|^2 = O(\mu),
\label{eq:qBarFundBound}
\eeq 
\beq 
\limsup_{i \rightarrow \infty} \E\|\widecheck{\bm{q}}_i\|^2 = O(\mu^2).
\label{eq:qCheckFundBound}
\eeq 

\section{Proof of Lemma~\ref{lem:simpModelApprox}}
\label{app:simpRecursmModelApprox}
\begin{IEEEproof}
Using the inverse network coordinate transformation in \eqref{jordan} we can write
\beq 
\widetilde{\bm{q}}_i = \mathds{1}_N \otimes \bar{\bm{q}}_i + (V_L \otimes I_M)\,\widecheck{\bm{q}}_i,
\eeq 
and applying definition \eqref{eq:qCov}:
\begin{align}
\Pi_i & = \mathds{1}_N \mathds{1}_N^{\top} \otimes \bar{\Pi}_i + \E\left[ (V_L \otimes I_M)\,\widecheck{\bm{q}}_i\widecheck{\bm{q}}_i^{\top} (V_L \otimes I_M)^{\top}\right] \nonumber \\
& + \E\left\{ (\mathds{1}_N \otimes \bar{\bm{q}}_i)\left[(V_L \otimes I_M)\,\widecheck{\bm{q}}_i\right]^{\top}\right\} \nonumber \\
& +  \E\left\{ \left[(V_L \otimes I_M)\,\widecheck{\bm{q}}_i\right](\mathds{1}_N \otimes \bar{\bm{q}}_i)^{\top}\right\},
\label{eq:piDef}
\end{align}
where
\beq 
\bar{\Pi}_i \triangleq \E\left[ \bar{\bm{q}}_i \bar{\bm{q}}_i^{\top}\right].
\label{eq:piBarDef}
\eeq 
Computing the norm of \eqref{eq:piDef} we have that
\begin{align}
& \| \Pi_i - \mathds{1}_N \mathds{1}_N^{\top} \otimes \bar{\Pi}_i\| \nonumber \\
& \overset{(a)}{\leq} \left\|\E\left[ (V_L \otimes I_M)\,\widecheck{\bm{q}}_i\widecheck{\bm{q}}_i^{\top} (V_L \otimes I_M)^{\top}\right]\right\| \nonumber \\
& + 2\left\|\E\left\{ (\mathds{1}_N \otimes \bar{\bm{q}}_i)\left[(V_L \otimes I_M)\,\widecheck{\bm{q}}_i\right]^{\top}\right\}\right\| \nonumber \\
& \overset{(b)}{\leq} \E\| (V_L \otimes I_M)\,\widecheck{\bm{q}}_i \|^2 \nonumber \\
& + 2\left\|\E\left\{ (\mathds{1}_N \otimes \bar{\bm{q}}_i)\left[(V_L \otimes I_M)\,\widecheck{\bm{q}}_i\right]^{\top}\right\}\right\| \nonumber \\
& \overset{(c)}{\leq} \|(V_L \otimes I_M)\|^2\E\|\widecheck{\bm{q}}_i\|^2 \nonumber \\
& + \sqrt{\E\|\mathds{1}_N \otimes \bar{\bm{q}}_i\|^2 \E\|(V_L \otimes I_M)\,\widecheck{\bm{q}}_i\|^2},
\label{eq:piDefDecomp}
\end{align}
where step $(a)$ follows from the triangle inequality, step $(b)$ follows from Jensen's inequality $\|\E[\cdot]\| \leq \E\|\cdot\|$ and the submultiplicative property of the norm, and step $(c)$ applies the Cauchy-Schwarz inequality. Applying the limit superior to \eqref{eq:piDefDecomp}, in view of the bounds \eqref{eq:qBarFundBound} and \eqref{eq:qCheckFundBound}, we conclude that 
\beq  
\limsup_{i \rightarrow \infty} \| \Pi_i - \mathds{1}_N \mathds{1}_N^{\top} \otimes \bar{\Pi}_i\| \leq O(\mu^{3/2}).
\label{eq:piPiBarClose}
\eeq 
We now move to prove that $\bar{\Pi}_i$ is equal to the correlation matrix $\Pi_{a,i}$ in \eqref{eq:qCovSimp} up to higher-order perturbations, which in turn implies that $\Pi_i$ is close to $\Pi_{a,i}$ since \eqref{eq:piPiBarClose} holds.
From the first row of \eqref{qTildeSystem} and the definition of $\bm{q}_{a,i}$ in \eqref{qaSystem}, we see that
\begin{align}
& \E\|\bar{\bm{q}}_{i} - \bm{q}_{a,i}\|^2 \nonumber \\ & = \E\|(I_M - \mu\,\zeta\, G_{11})(\bar{\bm{q}}_{i-1} - \bm{q}_{a,i-1}) - \mu\,\zeta\, G_{12}\,\widecheck{\bm{q}}_{i-1}\|^2 \nonumber \\ & 
\leq (1-\mu\,\zeta\,\nu)\E\|\bar{\bm{q}}_{i-1} - \bm{q}_{a,i-1}\|^2 + \frac{\mu\,\zeta\,\sigma_{12}^2}{\nu}\E\|\widecheck{\bm{q}}_{i-1}\|^2,
\label{eq:qiQaiDiffNormPre}
\end{align}
where the inequality comes from the application of Jensen's inequality with convex weights $1 - \mu\,\zeta\,\nu$ and $\mu\,\zeta\,\nu$, and the bounds \eqref{eq:IminusG11TimeBound} and (152) from \cite{CarpentieroMattaSayed2022}. Applying the limit superior to expression \eqref{eq:qiQaiDiffNormPre}
we conclude that
\begin{align}
& \limsup_{i \rightarrow \infty} \E\|\bar{\bm{q}}_{i} - \bm{q}_{a,i}\|^2 \nonumber \\
& \leq (1-\mu\,\zeta\,\nu) \limsup_{i \rightarrow \infty} \E\|\bar{\bm{q}}_{i} - \bm{q}_{a,i}\|^2 + O(\mu^3),
\label{eq:qiQaiDiffNorm}
\end{align}
where the first inequality follows from \eqref{eq:qCheckFundBound}. 
Moving to the LHS  the limit superior appearing on the RHS, from \eqref{eq:qiQaiDiffNorm} we obtain:
\beq
\limsup_{i \rightarrow \infty} \E\|\bar{\bm{q}}_{i} - \bm{q}_{a,i}\|^2=O(\mu^2).
\label{eq:qiQaiDiffNormfinalest}
\eeq
From the definition of the correlation matrix in \eqref{eq:piBarDef} we have
\begin{align}
\bar{\Pi}_{i} & = \E\left[ \bar{\bm{q}}_i \bar{\bm{q}}_i^{\top}\right] \nonumber \\
& = \E\left[ (\bm{q}_{a,i} + \bar{\bm{q}}_i - \bm{q}_{a,i})(\bm{q}_{a,i} + \bar{\bm{q}}_i - \bm{q}_{a,i})^{\top} \right] \nonumber \\
& = \E\left[ \bm{q}_{a,i}\bm{q}_{a,i}^{\top}\right] + \E\left[\bm{q}_{a,i}(\bar{\bm{q}}_i - \bm{q}_{a,i})^{\top} \right] \nonumber \\
& + \E\left[ (\bar{\bm{q}}_i - \bm{q}_{a,i})\bm{q}_{a,i}^{\top}\right] \!+\! \E\left[(\bar{\bm{q}}_i - \bm{q}_{a,i})(\bar{\bm{q}}_i - \bm{q}_{a,i})^{\top} \right] \nonumber \\
& = \Pi_{a,i} + \E\left[ \bar{\bm{q}}_i(\bar{\bm{q}}_i - \bm{q}_{a,i})^{\top}\right] + \E\left[ (\bar{\bm{q}}_i - \bm{q}_{a,i})\bar{\bm{q}}_i^{\top}\right] \nonumber \\
& -\E\left[ (\bar{\bm{q}}_i - \bm{q}_{a,i})(\bar{\bm{q}}_i - \bm{q}_{a,i})^{\top}\right],
\label{eq:splitLongNorm}
\end{align}
where in the last equality we used the definition of the correlation matrix in \eqref{eq:qCovSimp}. We now examine the steady-state behavior of the last three terms in \eqref{eq:splitLongNorm}:
\begin{align}
& \limsup_{i \rightarrow \infty} \left\| \E\left[ \bar{\bm{q}}_i(\bar{\bm{q}}_i - \bm{q}_{a,i})^{\top}\right] \right\| \nonumber \\
& \overset{(a)}{\leq} \limsup_{i \rightarrow \infty} \E\left\| \bar{\bm{q}}_i(\bar{\bm{q}}_i - \bm{q}_{a,i})^{\top} \right\| \nonumber \\
& \overset{(b)}{\leq} \limsup_{i \rightarrow \infty} \E\left[ \|\bar{\bm{q}}_i\|\,\|\bar{\bm{q}}_i - \bm{q}_{a,i} \|\right] \nonumber \\
& \overset{(c)}{\leq} \limsup_{i \rightarrow \infty} \sqrt{\E\|\bar{\bm{q}}_i\|^2 \E\|\bar{\bm{q}}_i - \bm{q}_{a,i}\|^2}  = O(\mu^{3/2}),
\label{eq:crossTerm1}
\end{align}
where step $(a)$ is an application of Jensen's inequality to the convex function $\|\cdot\|$; step $(b)$ follows from submultiplicativity of the operator norm; step $(c)$ uses the Cauchy-Schwarz inequality; and the last equality uses \eqref{eq:qBarFundBound} and \eqref{eq:qiQaiDiffNorm}. The same analysis holds also for the second term in \eqref{eq:splitLongNorm}. Finally, for the third term we can write: 
\begin{align}
&\limsup_{i \rightarrow \infty} \left\| \E\left[ (\bar{\bm{q}}_i - \bm{q}_{a,i})(\bar{\bm{q}}_i - \bm{q}_{a,i})^{\top}\right] \right\| \nonumber \\
& \leq \limsup_{i \rightarrow \infty} \E\left\| (\bar{\bm{q}}_i - \bm{q}_{a,i})(\bar{\bm{q}}_i - \bm{q}_{a,i})^{\top} \right\| \nonumber \\
& \leq \limsup_{i \rightarrow \infty} \E\| \bar{\bm{q}}_i - \bm{q}_{a,i} \|^2 = O(\mu^2),
\label{eq:crossTerm2}
\end{align}
where the first inequality is again an application of Jensen's inequality, the second inequality uses the submultiplicative property of the norm and the last step follows from  \eqref{eq:qiQaiDiffNormfinalest}. Computing the norm of \eqref{eq:splitLongNorm} and then using \eqref{eq:crossTerm1} and \eqref{eq:crossTerm2} we obtain
\beq 
\limsup_{i \rightarrow \infty} \| \bar{\Pi}_i - \Pi_{a,i}\| = O(\mu^{3/2}).
\label{eq:piBarPiAClose}
\eeq 
We can now write
\begin{align}
& \limsup_{i \rightarrow \infty} \|\Pi_i - \mathds{1}_N\mathds{1}_N^{\top} \otimes \Pi_{a,i} \| \nonumber \\
& = \limsup_{i \rightarrow \infty} \| \Pi_i - \mathds{1}_N\mathds{1}_N^{\top} \otimes\bar{\Pi}_{i} +   \mathds{1}_N\mathds{1}_N^{\top} \otimes(\bar{\Pi}_{i}  - \Pi_{a,i}) \| \nonumber \\
& \overset{(a)}{\leq} \limsup_{i \rightarrow \infty} \|\Pi_i - \mathds{1}_N\mathds{1}_N^{\top} \otimes\bar {\Pi}_{i}\| \nonumber \\
& + \limsup_{i \rightarrow \infty} \| \mathds{1}_N\mathds{1}_N^{\top} \otimes(\bar{\Pi}_{i}  - \Pi_{a,i}) \| \nonumber \\
& \overset{(b)}{=} \limsup_{i \rightarrow \infty} \|\Pi_i - \mathds{1}_N\mathds{1}_N^{\top} \otimes\bar{\Pi}_{i}\| \nonumber \\ & + \limsup_{i \rightarrow \infty} \|\mathds{1}_N\mathds{1}_N^{\top} \| \|\bar{\Pi}_{i}  - \Pi_{a,i}\| =  O(\mu^{3/2}),
\label{eq:covaDecomp}
\end{align}
where step $(a)$ follows from the triangle inequality, 
step $(b)$ follows from the fact that the $2$-induced matrix norm is equal to the largest singular value and the singular values of a Kronecker product $X \otimes Y$ are given by the products of the singular values of $X$ and $Y$, for any two suitable matrices $X$ and $Y$. The final equality uses \eqref{eq:piPiBarClose} and \eqref{eq:piBarPiAClose}.

Introducing the notation:
\beq
\widetilde{\mathcal{R}}_{i-1}\triangleq \E\left[\mathcal{R}_{s}(\bm{w}_{i-1})\right] - \mathcal{R}_{s}(\mathds{1}_N\otimes w^o),
\eeq
from \eqref{eq:covComp2} and \eqref{eq:covComp3} the difference between $\Pi_{a,i}$ and $\Pi_{a,i}^o$ is upper bounded by
\begin{align} 
& \|\Pi_{a,i} - \Pi_{a,i}^o\| \nonumber \\
& \leq \|D\|^2\|(\Pi_{a,i-1} - \Pi_{a,i-1}^o)\| 
\nonumber\\
&+ \left\|\mu^2\zeta^2\left(\pi^{\top} \otimes I_M\right)
\widetilde{\mathcal{R}}_{i-1}
\left(\pi \otimes I_M\right)\right\| \nonumber \\
& \leq (1-\mu\,\zeta\,\nu)^2\|(\Pi_{a,i-1} - \Pi_{a,i-1}^o)\| \nonumber \\
& + \left\|\mu^2\zeta^2\left(\pi^{\top} \otimes I_M\right)
\widetilde{\mathcal{R}}_{i-1}
\left(\pi \otimes I_M\right)\right\|,
\label{eq:unperturbedComp}
\end{align}
where the first inequality follows from the triangle inequality, and in the second step we used \eqref{eq:IminusG11TimeBound}. 
The norm of the last term on the RHS of \eqref{eq:unperturbedComp} can be decomposed as follows:
\begin{align}
& \left\|\mu^2\zeta^2(\pi^{\top} \otimes I_M)
\widetilde{\mathcal{R}}_{i-1}
\left(\pi \otimes I_M\right)\right\| \nonumber \\
& \leq \mu^2\zeta^2\|\pi^{\top} \otimes I_M\|
\times\|\widetilde{\mathcal{R}}_{i-1}\|
\times\|\pi \otimes I_M\| \nonumber \\
& \overset{(a)}{=} \mu^2\zeta^2\| \pi\|^2\|
\widetilde{\mathcal{R}}_{i-1}
\| \nonumber \\
& \overset{(b)}{=} \mu^2\zeta^2\|\pi\|^2 \max_{1 \leq k \leq N} \|\E\left[R_{s,k}(\bm{w}_{k,i-1}) - R_{s,k}(w^o)\right]\| \nonumber \\
& \overset{(c)}{\leq} \mu^2\zeta^2\|\pi\|^2 \max_{1 \leq k \leq N} \E\|R_{s,k}(\bm{w}_{k,i-1}) - R_{s,k}(w^o)\|,
\label{eq:higherOrderPertDecomp}
\end{align}
where step $(a)$ follows from the fact that the singular values of $A \otimes I_M$ are the (replicated) singular values of $A$, for any matrix $A$, step $(b)$ exploits the diagonal shape of $\mathcal{R}_{s}(\bm{w}_{i-1})$ and $\mathcal{R}_{s}(\mathds{1}_N\otimes w^o)$, and step $(c)$ is an application of Jensen's inequality to the convex function $\|\cdot\|$.
From \eqref{eq:covdifflimsupmain} we conclude that:
\beq
\limsup_{i \rightarrow \infty} \left\|\mu^2\zeta^2(\pi^{\top} \otimes I_M)\widetilde{\mathcal{R}}_{i-1}\left(\pi \otimes I_M\right)\right\| 
= O(\mu^{3}).
\label{eq:perturbOrder}
\eeq
Applying the limit superior to \eqref{eq:unperturbedComp}, and then \eqref{eq:perturbOrder}, gives
\beq 
\limsup_{i \rightarrow \infty} \|\Pi_{a,i} - \Pi_{a,i}^o\| = O(\mu^{2 }).
\label{eq:covDiffNorm}
\eeq 
Exploiting the same chain of inequalities employed to obtain \eqref{eq:covaDecomp}, but using $\Pi_{a,i}^o$ and $\Pi_{a,i}$ in place of $\Pi_{a,i}$ and $\bar{\Pi}_{i}$, invoking \eqref{eq:covaDecomp} and \eqref{eq:covDiffNorm} one obtains \eqref{eq:simpModelApprox}.

It remains to prove \eqref{eq:scalarprodapproxappendix}.
We first evaluate the absolute difference between the interaction among the compressed states and the approximate model in \eqref{eq:covComp3}:
\begin{align}
& \left| \E\left[\widetilde{\bm{q}}_{\ell,i}^{\top}\widetilde{\bm{q}}_{k,i}\right] - \textnormal{Tr}\left[\Pi_{a,i}^o\right]\right| \nonumber \\
& \overset{(a)}{=} \Big| \textnormal{Tr}\left[\Pi_i(E_{\ell k} \otimes I_M)\right] \nonumber \\
& - \textnormal{Tr}\left[(\mathds{1}_N\mathds{1}_N^{\top} \otimes \Pi_{a,i}^o) (E_{\ell k} \otimes I_M)\right] \Big| \nonumber \\
& = \left| \textnormal{Tr}\left[ (\Pi_i - \mathds{1}_N\mathds{1}_N^{\top} \otimes \Pi_{a,i}^o)(E_{\ell k} \otimes I_M)\right] \right| \nonumber \\
& \overset{(b)}{\leq} 
\sum_{j=1}^{MN} \sigma_j(\Pi_i - \mathds{1}_N\mathds{1}_N^{\top} \otimes \Pi_{a,i}^o)\,
\sigma_j(E_{\ell k} \otimes I_M)
\nonumber\\
& \overset{(c)}{\leq} \left\|\Pi_i - \mathds{1}_N\mathds{1}_N^{\top} \otimes \Pi_{a,i}^o\right\|
\,\underbrace{\sum_{j=1}^{MN} \sigma_j(E_{\ell k} \otimes I_M)}_{=M},
\label{eq:innProdTrick}
\end{align}
where step $(a)$ extracts the quantity $\E[\widetilde{\bm{q}}_{\ell,i}^{\top}\widetilde{\bm{q}}_{k,i}]$ from the correlation matrix $\Pi_i$ as shown in \eqref{eq:selectionMatrix}, and exploits the equality:
\beq
\textnormal{Tr}\left[\Pi_{a,i}^o\right]=\textnormal{Tr}\left[(\mathds{1}_N\mathds{1}_N^{\top} \otimes \Pi_{a,i}^o) (E_{\ell k} \otimes I_M) \right].
\eeq
Step $(b)$ uses Von Neumann's trace inequality~\cite{Bhatia}, where $\sigma_j(X)$ denotes the $j$-th singular value of matrix $X$. 
Finally, step $(c)$ applies the definition of $\ell_2$ induced matrix norm and the fact that the matrix $E_{\ell k} \otimes I_M$ has only $M$ nonzero singular values all equal to $1$.\footnote{The singular values of $E_{\ell k} \otimes I_M$ are computed by observing that $(E_{\ell k} \otimes I_M)(E_{\ell k} \otimes I_M)^{\top}=E_{\ell k} E_{k\ell}\otimes I_M$, and that the matrix $E_{\ell k} E_{k\ell}$ has all zero entries, but for the $(\ell,\ell)$ entry that is equal to $1$.}
Applying the limit superior to \eqref{eq:innProdTrick}, from Lemma~\ref{lem:simpModelApprox} we obtain \eqref{eq:scalarprodapproxappendix}. 
\end{IEEEproof}

\section{Proof of Lemma~\ref{lem:deltaApprox}} \label{app:weightedDiff}
\begin{IEEEproof}
Consider the evolution of the transformed differential iterates $\bm{\delta}_i$, which can be obtained by applying the inverse transformation $\mathcal{V}^{-1}$ to the evolution of transformed iterates $\widehat{\bm{\delta}}_i$ in \eqref{eq:otherDeltaTransRep}, namely,
\beq  
\bm{\delta}_i = \mathcal{V}^{-1}\underbrace{\left( \mathcal{J}_{\rm{tot}} - I_{MN} - \mu\,\mathcal{G}\right)}_{\mathcal{T}}\,\widehat{\bm{q}}_{i-1} - \mu\,\bm{s}_i.
\label{eq:deltaFullExpr}
\eeq  
Applying the $\Omega_{\pi}$ weighted norm and taking expectation in \eqref{eq:deltaFullExpr} we have 
\beq
\E\|\bm{\delta}_i\|^2_{\Omega_{\pi}}  = \E\|\mathcal{V}^{-1}\mathcal{T}\,\widehat{\bm{q}}_{i-1} \|^2_{\Omega_{\pi}} + \mu^2\E\|\bm{s}_i\|^2_{\Omega_{\pi}}, 
\label{eq:deltaWighNorm}
\eeq
where the equality follows from the zero-mean property of the gradient noise process.

Exploiting the structure of the matrices $\mathcal{V}^{-1}$ and $\mathcal{T}$, which are shown in \eqref{jordan} and \eqref{eq:deltaSystem}, respectively, we have that 
\beq
\mathcal{V}^{-1}\mathcal{T} = \begin{bmatrix}
-\mu\,\bar{G} & \widecheck{G}_1 - \mu\,\widecheck{G}_2,
\label{eq:TmatDef}
\end{bmatrix},
\eeq
where we defined 
\begin{align}
\bar{G} & \triangleq (\mathds{1}_N \otimes I_M)\, G_{11} + (V_L \otimes I_M )\, G_{21}, \\
\widecheck{G}_1 & \triangleq (V_L \otimes I_M)(\mathcal{J} - I_{M(N-1)}), \\
\widecheck{G}_2 & \triangleq (\mathds{1}_N \otimes I_M)\, G_{12} + (V_L \otimes I_M)\,G_{22}.
\end{align}   
Substituting \eqref{eq:TmatDef} into \eqref{eq:deltaWighNorm}, and recalling from \eqref{qTildeSystem} that $\widehat{\bm{q}}_i = \textnormal{col}\{ \bar{\bm{q}}_i, \widecheck{\bm{q}}_i\}$, we can write
\begin{align}
& \E\|\bm{\delta}_i\|^2_{\Omega_{\pi}} \nonumber \\
& \overset{(a)}{\leq} 2\,\mu^2\E\|\bar{G}\bar{\bm{q}}_{i-1}\|^2_{\Omega_{\pi}}  +  2\,\E\|(\widecheck{G}_1 - \mu\,\widecheck{G}_2)\,\widecheck{\bm{q}}_{i-1}\|^2_{\Omega_{\pi}} \nonumber \\
&  + \mu^2\E\|\bm{s}_i\|^2_{\Omega_{\pi}} \nonumber \\
& \overset{(b)}{\leq} 
2\,\mu^2\E\|\bar{G}\bar{\bm{q}}_{i-1}\|^2_{\Omega_{\pi}}  
+ 4\,\E\|\widecheck{G}_1\widecheck{\bm{q}}_{i-1}\|^2_{\Omega_{\pi}} \nonumber \\
& + 4\,\mu^2\E\|\widecheck{G}_2\widecheck{\bm{q}}_{i-1}\|^2_{\Omega_{\pi}}  + \mu^2\E\|\bm{s}_i\|^2_{\Omega_{\pi}} \nonumber \\
& \overset{(c)}{=} 
2\,\mu^2\|\Omega_{\pi}^{1/2}\bar{G}\|^2\E\|\bar{\bm{q}}_{i-1} \|^2 + 4\,\|\Omega_{\pi}^{1/2}\widecheck{G}_1\|^2\|\widecheck{\bm{q}}_{i-1}\|^2 \nonumber \\
& + 4\,\mu^2\|\Omega_{\pi}^{1/2}\widecheck{G}_2\|^2\|\widecheck{\bm{q}}_{i-1}\|^2  + \mu^2\E\|\bm{s}_i\|^2_{\Omega_{\pi}},
\label{eq:furtherDeltaSplitWeighting}
\end{align}
where $(a)$ and $(b)$ follow by repeated application of Jensen's inequality with uniform weights equal to $1/2$. For step $(c)$, note that for any positive definite matrix $W$ we can write
\begin{align}   
\|A x\|_W^2 & = x^{\top}A^{\top}W A x 
 = x^{\top}A^{\top}W^{1/2} W^{1/2} A \, x \nonumber \\
& = \|W^{1/2}Ax\|^2 \leq \|W^{1/2}A\|^2\,\|x\|^2,
\end{align}
which holds for any matrix $A$ and vector $x$ of compatible sizes. 
Taking the limit superior in \eqref{eq:furtherDeltaSplitWeighting} we have that 
\begin{align}  
\limsup_{i \rightarrow \infty} \E\|\bm{\delta}_i\|^2_{\Omega_{\pi}} & \leq \limsup_{i \rightarrow \infty} 4\,\|\Omega_{\pi}^{1/2} \widecheck{G}_1\|^2\E\|\widecheck{\bm{q}}_{i-1}\|^2 \nonumber \\
& +\!  \limsup_{i \rightarrow \infty} \mu^2\sum_{k=1}^N \pi_k^2\omega_k\E\|\bm{s}_{k,i}\|^2  \!+\! O(\mu^3),
\label{eq:deltaSupBound}
\end{align} 
where we used \eqref{eq:qBarFundBound} and \eqref{eq:qCheckFundBound}, along with the fact that the norms of the matrices $G_{11}$, $G_{12}$, $G_{21}$, and $G_{22}$, required to evaluate $\|\Omega_{\pi}^{1/2}\bar{G}\|$ and $\|\Omega_{\pi}^{1/2}\widecheck{G}_2\|$, can be bounded by constants as shown in \cite{Sayed,CarpentieroMattaSayed2022}. Moreover, we wrote the weighted norm of the gradient noise as a sum of individual norms because of the diagonal structure of $\Omega_{\pi}$.

To examine the first term on the RHS of \eqref{eq:deltaSupBound} we can use the characterization \cite{CarpentieroMattaSayed2022}:
\beq 
\limsup_{i \rightarrow \infty} \E\|\widecheck{\bm{q}}_i\|^2 \leq \mu^2\zeta\,c',
\label{eq:asymQCheckOld}
\eeq 
where $c'$ is a suitable constant independent of $\mu$, but whose value depends on the compression operator parameters $\{\omega_k\}_{k=1}^N$. The exact definition of $c'$ can be obtained from the non-asymptotic bound of the transformed compressed iterate $\widehat{\bm{q}}_{i}$ provided in \cite{CarpentieroMattaSayed2022}, in particular see Appendix~J therein. Substituting \eqref{eq:asymQCheckOld} in \eqref{eq:deltaSupBound} we have
\begin{align}  
\limsup_{i \rightarrow \infty} \E\|\bm{\delta}_i\|^2_{\Omega_{\pi}} & \leq \limsup_{i \rightarrow \infty} \mu^2\sum_{k=1}^N \pi_k^2\omega_k\E\|\bm{s}_{k,i}\|^2 \nonumber \\
& + \mu^2\zeta\,c + O(\mu^3),
\label{eq:deltaSupBoundAfterQcheck}
\end{align} 
where $c \triangleq 4\,\|\Omega_{\pi}^{1/2} \widecheck{G}_1\|^2c'$. 

Using \eqref{eq:meansquaregradmaintheorem} in \eqref{eq:deltaSupBoundAfterQcheck} completes the proof.
\end{IEEEproof}

\section{Auxiliary Lemma}
\label{app:geomSum}
\begin{lemma}[{\bf Useful geometric summation}]\label{lem:geomSum}
Let $\beta \in (0,1)$. 
Consider a sequence $\{z_i\}$ and let:
\beq 
\limsup_{i \rightarrow \infty} z_i = z <\infty.
\eeq
Then, it holds that
\beq 
\limsup_{i\rightarrow \infty} \sum_{j=0}^i \beta^j z_{i-j} \leq \frac{z}{1 - \beta}.
\label{eq:geomSumPiezzo}
\eeq 
\end{lemma}
\begin{IEEEproof}
By definition of limit superior, for an arbitrary $\epsilon > 0$ there exists an index $i_0(\epsilon)$ such that $z_i < z + \epsilon$ for all $i \geq i_0(\epsilon)$. 
Thus, we can write
\begin{align}
\sum_{j=0}^i \beta^j z_{i-j} & = \sum_{j=0}^{i - i_0(\epsilon)}\beta^j z_{i-j} + \sum_{j = i - i_0(\epsilon)+1}^i \beta^j z_{i-j} \nonumber \\
& < \left(z + \epsilon\right)\sum_{j=0}^{i - i_0(\epsilon)}\beta^j  + i_0(\epsilon)\,z_{\sup}\, \beta^{i - i_0(\epsilon) + 1},
\label{eq:geomSumProofStep1}
\end{align} 
where we denoted by $z_{\sup}$ the supremum of the bounded sequence $\{z_i\}$. 
Taking the limit superior on both sides of \eqref{eq:geomSumProofStep1} we obtain
\begin{align}
&\limsup_{i \rightarrow \infty} \sum_{j=0}^i \beta^j z_{i-j} \leq (z+\epsilon)\,\limsup_{i \rightarrow \infty}  
\underbrace{\sum_{j=0}^{i - i_0(\epsilon)}\beta^j}_{\leq \frac{1}{1-\beta}} \nonumber \\
&  + i_0(\epsilon)\,z_{\sup}\, \underbrace{\limsup_{i \rightarrow \infty} \beta^{i - i_0(\epsilon)+1}}_{=0}
\leq z+\epsilon,
\end{align}
and the claim follows from the arbitrariness of $\epsilon$.
\end{IEEEproof}

\section{Solution to the Resource Allocation Problem}
\label{app:KKT}
\subsection{Optimized Allocation for the Randomized Quantizers}
We first consider the resource allocation problem when employing the randomized quantizers, i.e., using \eqref{eq:approxCompAlistarh} in \eqref{eq:secondOptProblem}. In this case, the domain of the optimization problem is given by $x \in \mathbb{R}^N$ such that $x \succ 0$. Note that the objective function is strictly convex and differentiable in this set. 
The Lagrangian associated with the objective function is equal to
\begin{align} 
\mathcal{L}(x,\lambda) & = \sum_{k=1}^N \pi_k^2 d_k \frac{M}{(2^{x_k} - 1)^2}  + \sum_{k=1}^N\lambda_k^-(x_{\rm{min}} - x_k) \nonumber \\
& + \sum_{k=1}^N\lambda_k^+(x_k - x_{\rm{max}}) + \lambda_0\left(\sum_{k=1}^N x_k - X\right),
\end{align} 
where $\lambda = [\lambda_0, \lambda_1^-,\lambda_2^-,\ldots,\lambda_N^-, \lambda_1^+,\lambda_2^+,\ldots,\lambda_N^+] \succeq 0$ are the Lagrange multipliers related to the constraints of the problem, and the quantity $d_k$ is defined in \eqref{eq:geomMeanPerron}.

The KKT conditions associated with \eqref{eq:secondOptProblem}, for $k=1,2,\ldots,N$ are~\cite{Boyd-Vandenberghe}:
\begin{align}
& \frac{\partial \mathcal{L}(x,\lambda)}{\partial x_k} = -\frac{M\log(2)\pi^2_k d_k 2^{x_k+1}}{(2^{x_k}-1)^3} \!-\! \lambda_k^- \!+\! \lambda_k^+ \!+\! \lambda_0 = 0,
\label{eq:statCond}
\\
& \lambda_k^-(x_{\rm{min}} - x_k) = 0, \qquad \lambda_k^+(x_k - x_{\rm{max}}) = 0,
\label{eq:compSlack12}  
\\
& x_{\rm{min}} - x_k \leq 0, \qquad  x_k - x_{\rm{max}} \leq 0, 
\label{eq:xminConst}
\\
& \lambda_k^-, \lambda_k^+ \geq 0, 
\label{eq:lastKKTCond}
\end{align}
to which we must add three more conditions that hold irrespective of $k$, namely,
\begin{align}
& \lambda_0\left(\sum_{k=1}^N x_k - X\right) = 0, 
\label{eq:compSlack3}
\\
& \sum_{k=1}^N x_k - X \leq 0, 
\label{eq:ineqConstX}
\\
& \lambda_0 \geq 0.
\label{eq:veryLastKKTCond}
\end{align}
To find a solution $(x, \lambda)$ that satisfies the KKT conditions we consider two separate cases, starting from condition \eqref{eq:compSlack3}. 

In the first one we have that $\sum_{k=1}^N x_k = X$, which allows us to set $\lambda_0 > 0$. We then have three possible rules to find the unknown allocation $x_k$, which derive from conditions \eqref{eq:compSlack12}:
\begin{itemize}
\item [$i)$] $\lambda_k^- = \lambda_k^+ = 0$ and therefore $x_k$ is the solution to the equation\footnote{Note that the function $2^{x}/(2^{x}-1)^3$ is monotonically decreasing for $x > 0$ and tends asymptotically to zero. Therefore, equation \eqref{eq:funnyEquation1} has one, and only one solution for $x > 0$.}
\beq 
\lambda_0 = \frac{M\log(2)\pi^2_k d_k 2^{x_k+1}}{(2^{x_k}-1)^3}.
\label{eq:funnyEquation1}
\eeq
\item [$ii)$] $\lambda_k^- = 0$, $x_k = x_{\rm{max}}$, yielding 
\beq  
\lambda_k^+ = \frac{M\log(2)\pi^2_k d_k 2^{x_{\rm{max}}+1}}{(2^{x_{\rm{max}}}-1)^3}  - \lambda_0.
\label{eq:funnyEquation2}
\eeq 
\item [$iii)$] $\lambda_k^+ = 0$, $x_k = x_{\rm{min}}$, yielding
\beq   
\lambda_k^- = \lambda_0 - \frac{M\log(2)\pi^2_k d_k 2^{x_{\rm{min}}+1}}{(2^{x_{\rm{min}}}-1)^3}.
\label{eq:funnyEquation3}
\eeq 
\end{itemize}

In the second case, if $\sum_{k=1}^N x_k < X$, we have to set $\lambda_0 = 0$ for \eqref{eq:compSlack3} to hold. Repeating the same arguments that lead to \eqref{eq:funnyEquation1}-\eqref{eq:funnyEquation3}, the only admissible condition is
$\lambda_k^- = 0$, $x_k = x_{\rm{max}}$ and
\beq   
\lambda_k^+ = \frac{M\log(2)\pi^2_k d_k 2^{x_{\rm{max}}+1}}{(2^{x_{\rm{max}}}-1)^3}.
\label{eq:funnyEquation3}
\eeq 
The minimizer $x$ is obtained following the rules \eqref{eq:funnyEquation1} to \eqref{eq:funnyEquation3} after choosing, numerically, the value for the multiplier $\lambda_0 > 0$ such that the remaining constraints \eqref{eq:xminConst} to \eqref{eq:lastKKTCond} are verified.

\subsection{Optimized Allocation for the Randomized Sparsifiers}
The allocation problem when using the randomized sparsifiers can be solved applying again the KKT conditions, since the objective function using the compression parameter in \eqref{eq:randSparsOmega} is strictly convex and differentiable in the domain $x \in \mathbb{R}^N$ such that $x \succ 0$. In this case the Lagrangian is equal to
\begin{align} 
\mathcal{L}(x,\lambda) & = \sum_{k=1}^N \pi_k^2 d_k \left(\frac{M}{x_k} - 1 \right) + \sum_{k=1}^N\lambda_k^-(x_{\rm{min}} - x_k) \nonumber \\
& + \sum_{k=1}^N\lambda_k^+(x_k - x_{\rm{max}}) + \lambda_0\left(\sum_{k=1}^N x_k - X\right).
\end{align} 
We obtain the same conditions \eqref{eq:statCond}-\eqref{eq:veryLastKKTCond}, where condition \eqref{eq:statCond} takes the form
\beq
\frac{\partial \mathcal{L}(x,\lambda)}{\partial x_k} = -\frac{M\pi_k^2d_k }{x_k^2} - \lambda_k^- + \lambda_k^+ + \lambda_0 = 0, \quad  \forall k. 
\label{eq:statCondSpars}
\eeq
Using the same approach applied to solve \eqref{eq:statCond}-\eqref{eq:lastKKTCond} when $\sum_{k=1}^N x_k = X$ and $\lambda_0 > 0$, we end up with the following rules
\begin{itemize}
\item [$i)$] $\lambda_k^- = \lambda_k^+ = 0\Rightarrow x_k = \sqrt{\left( M\pi^2_k d_k\right)/\lambda_0}$ 
\item [$ii)$] $\lambda_k^- = 0$, $x_k = x_{\rm{max}}\Rightarrow \lambda_k^+ = \left(M\pi^2_k d_k\right)/x_{\rm{max}}^2 - \lambda_0$. 
\item [$iii)$] $\lambda_k^+ = 0$, $x_k = x_{\rm{min}}$, yielding
\beq  
\lambda_k^- = \lambda_0 - \left(M\pi^2_k d_k\right)/x_{\rm{min}}^2.
\eeq 
\end{itemize}
In the case $\sum_{k=1}^N x_k < X$ and $\lambda_0 = 0$ we get the condition
$\lambda_k^- = 0$, $x_k = x_{\rm{max}}$ and 
\beq   
\lambda_k^+ = \left(M\pi^2_k d_k\right)/x_{\rm{max}}^2.
\eeq  
Choosing, numerically, the value for the multiplier $\lambda_0 > 0$ that allows to satisfy the constraints \eqref{eq:xminConst} to \eqref{eq:lastKKTCond} gives the minimizer $x$.


\end{document}